\documentclass[runningheads]{llncs}

\usepackage{eccv}

\usepackage{eccvabbrv}

\usepackage{graphicx}
\usepackage{booktabs}
\usepackage{multirow}
\usepackage{threeparttable}
\usepackage{hyperref}
\usepackage{url}
\usepackage{pifont}
\usepackage{colortbl}
\usepackage{makecell}
\definecolor{softblue}{rgb}{0.8, 0.9, 1.0}
\usepackage{pifont}
\usepackage{mathrsfs}
\usepackage{wrapfig}
\usepackage{caption}

\usepackage{adjustbox}
\usepackage{bbding}
\usepackage{orcidlink}
\usepackage{cite}

\usepackage[accsupp]{axessibility}  

\newcommand{\xmark}{\ding{55}}

\usepackage{hyperref}

\usepackage{orcidlink}

\begin{document}

\title{GeoSolver: Scaling Test-Time Reasoning in Remote Sensing with Fine-Grained Process Supervision} 

\titlerunning{ }


\author{
Lang Sun\inst{1,2} \and
Ronghao Fu\inst{1,2}\Envelope \and
Zhuoran Duan\inst{1,2} \and Haoran Liu\inst{1,2} \and Xueyan Liu\inst{1,2} \and Bo Yang\inst{1,2}\Envelope
}

\authorrunning{Sun et al.}


\institute{
$^1$College of Computer Science and Technology, Jilin University, Changchun 130012, China\\
$^2$Key Laboratory of Symbolic Computation and Knowledge Engineering of Ministry of Education Jilin University \\
\email{\{sunlang24,duanzr24,lhr24\}@mails.jlu.edu.cn, \{furh,xueyanliu,ybo\}@jlu.edu.cn}}
\def\customsymbol#1{
    \ifcase\number\value{#1}
        \or\Envelope
    \else\@ctrerr
    \fi
}

\maketitle
\renewcommand{\footnotesize}{\fontsize{8pt}{8pt}\selectfont}

\footnotetext[1]{Corresponding author.}

\begin{abstract}
While Vision-Language Models (VLMs) have significantly advanced remote sensing interpretation, enabling them to perform complex, step-by-step reasoning remains highly challenging. Recent efforts to introduce Chain-of-Thought (CoT) reasoning to this domain have shown promise, yet ensuring the visual faithfulness of these intermediate steps remains a critical bottleneck. To address this, we introduce GeoSolver, a novel framework that transitions remote sensing reasoning toward verifiable, process-supervised reinforcement learning. We first construct Geo-PRM-2M, a large-scale, token-level process supervision dataset synthesized via entropy-guided Monte Carlo Tree Search (MCTS) and targeted visual hallucination injection. Building upon this dataset, we train GeoPRM, a token-level process reward model (PRM) that provides granular faithfulness feedback. To effectively leverage these verification signals, we propose Process-Aware Tree-GRPO, a reinforcement learning algorithm that integrates tree-structured exploration with a faithfulness-weighted reward mechanism to precisely assign credit to intermediate steps. Extensive experiments demonstrate that our resulting model, GeoSolver-9B, achieves state-of-the-art performance across diverse remote sensing benchmarks. Crucially, GeoPRM unlocks robust Test-Time Scaling (TTS). Serving as a universal geospatial verifier, it seamlessly scales the performance of GeoSolver-9B and directly enhances general-purpose VLMs, highlighting its remarkable cross-model generalization. The code will be available at \url{https://github.com/minglangL/GeoSolver}.

\end{abstract}

\section{Introduction}
\label{sec:intro}

The integration of Vision-Language Models (VLMs) has profoundly advanced the field of remote sensing~\cite{kuckreja2024geochat, pang2025vhm, zhang2024earthgpt, soni2025earthdial, liu2025skymoe, liu2025geodit, guo2024skysense, Zhu_2025_CVPR_skysense_o}, transitioning systems from simple discriminative perception to complex, open-ended visual understanding. Recently, pioneering efforts have sought to elevate these models further by introducing the Chain-of-Thought (CoT) paradigm to the geospatial domain~\cite{li2025segearth, yao2025remotereasoner, fiaz2025geovlmr1, zhang2025geor1_fewshot}. By forcing models to decompose tasks and explicitly gather visual evidence before predicting an answer~\cite{liu2025towardsrsthinker, shao2025asking_geoeot, wang2025geozero}, these perceptually-grounded reasoning frameworks have demonstrated remarkable potential in interpreting dense, top-down satellite imagery.

\begin{figure}[tb]
    \centering
    \includegraphics[width=0.98\linewidth]{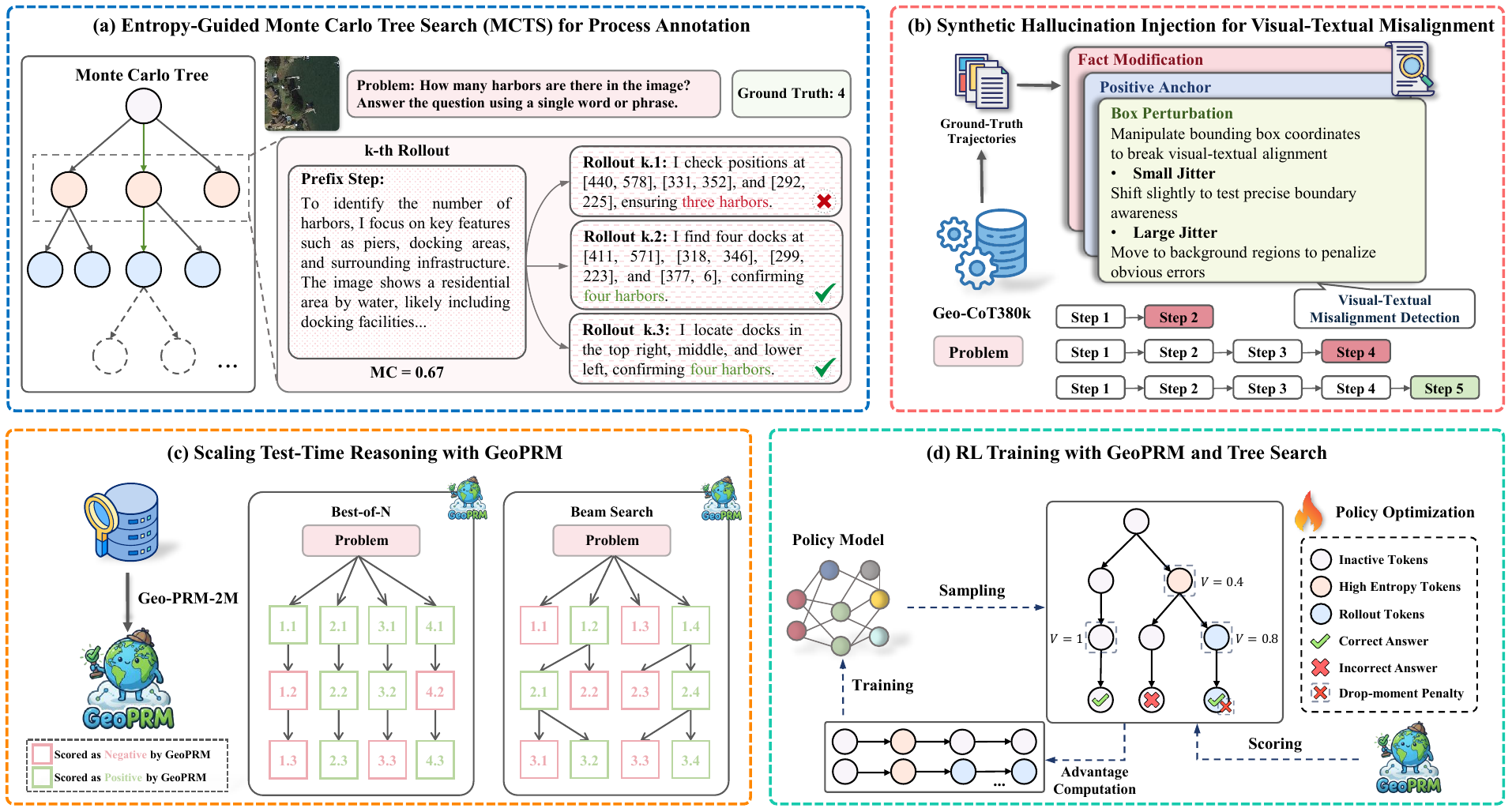}
    \vspace{-0.2cm}
    \caption{\textbf{Overall framework of our proposed GeoSolver.} To enable reliable geospatial reasoning, we first construct the Geo-PRM-2M dataset via (a) Entropy-Guided MCTS and (b) Synthetic Hallucination Injection. The trained GeoPRM is then seamlessly integrated to (c) scale test-time reasoning via advanced search strategies during inference, and (d) align the base policy via Process-Aware Tree-GRPO during training.}
    \label{fig:framework}
    \vspace{-0.3cm}
\end{figure}

However, optimizing these multimodal reasoning processes presents significant challenges. While traditional end-to-end remote sensing VLMs relied exclusively on Supervised Fine-Tuning (SFT)~\cite{kuckreja2024geochat, luo2024skysensegpt, shabbir2025geopixel}, recent reasoning-centric models have advanced by adopting Reinforcement Learning (RL) frameworks~\cite{liu2025towardsrsthinker, fiaz2025geovlmr1, wang2025geozero}, most notably Group Relative Policy Optimization (GRPO)~\cite{deepseekai2025deepseekr1incentivizingreasoningcapability}. Yet, current GRPO paradigms in this domain rely predominantly on outcome-based rewards. In visually complex geospatial scenarios, optimizing solely on final outcomes exacerbates the credit assignment problem. Without explicit step-level verification, models are frequently rewarded for "lucky guesses," where they generate linguistically fluent but visually ungrounded claims (e.g., hallucinating an incorrect bounding box that coincidentally leads to the right object count). Consequently, outcome-supervised policies are inadvertently encouraged to memorize spurious correlations rather than perform faithful visual grounding~\cite{xie2025outcomes_prms, peng2024multimath}. To eliminate these spatial hallucinations, transitioning to process supervision is imperative. While Process Reward Models (PRMs) have shown tremendous promise in domains like mathematical reasoning~\cite{huang2025visionr1_math, yang2024qwen2_5math, yu2025chainmath, zhuang2025mathpuma}, their application in remote sensing remains entirely uncharted. Furthermore, naively importing scalar PRMs as direct optimization targets is highly problematic, as it frequently induces length bias and reward hacking, driving the policy to generate artificially truncated reasoning to avoid step-wise penalties~\cite{luo2025ursa, peng2025rewarding_graphprm}.

To overcome these fundamental challenges, we present the \textbf{GeoSolver} framework, a novel approach that aligns remote sensing VLMs towards verifiable, step-by-step reasoning. Recognizing that robust process supervision requires domain-specific discriminative signals, we first construct \textbf{Geo-PRM-2M}, a large-scale token-level process supervision dataset. We synthesize this corpus through a dual-view pipeline: we employ an entropy-guided Monte Carlo Tree Search~\cite{browne2012survey_mcts} to mine the model's intrinsic logical errors and inject synthetic perturbations (e.g., bounding box jittering) to explicitly target visual grounding failures. Trained on this dataset, our token-level verifier, \textbf{GeoPRM}, provides granular, continuous feedback on reasoning faithfulness.

Building upon GeoPRM, we propose a novel RL alignment algorithm: \textbf{Process-Aware Tree-GRPO}. Rather than relying on sample-inefficient linear rollouts~\cite{shao2024deepseekmath}, we construct an entropy-guided reasoning tree during the RL exploration phase. Crucially, we introduce a faithfulness-weighted reward mechanism that penalizes trajectories exhibiting sudden drops in PRM confidence~\cite{luo2025ursa}, even if their final outcome is correct. By propagating these process-aware signals through the tree via global and local advantages, our algorithm explicitly incentivizes reasoning paths that are not only accurate but consistently verifiable.

Extensive experiments across six major remote sensing tasks demonstrate the efficacy of our approach. Applying our framework to train the GeoSolver-9B model, we show that under standard inference, it significantly outperforms both dedicated remote sensing models and general-purpose reasoning VLMs. Furthermore, by employing GeoPRM for test-time scaling (TTS)~\cite{muennighoff2025s1_tts, zhang2024generative_tts}, we observe consistent performance gains, validating the compute-optimal scaling law in remote sensing. Most notably, when used to guide general-purpose models, GeoPRM enables them to surpass fully fine-tuned domain experts, demonstrating that our reward model captures a generalized, fundamental logic of multimodal geospatial verification. All datasets, models, and code will be open-sourced upon publication.

In summary, our contributions can be summarized as follows:
\begin{itemize}
\item We construct Geo-PRM-2M, the first large-scale process supervision dataset for remote sensing, and develop GeoPRM, a token-level PRM that precisely localizes logical and visual hallucinations.
\item We propose Process-Aware Tree-GRPO, an RL algorithm that organically integrates efficient tree-structured exploration with step-wise verification, resolving the credit assignment inherent in standard reasoning alignment.
\item Experimental results show that GeoSolver-9B achieves leading performance across diverse geospatial tasks. Furthermore, we comprehensively validate the efficacy of Test-Time Scaling in remote sensing and showcase the remarkable cross-model generalization of our proposed PRM.
\end{itemize}

\section{Related Work}
\label{sec:related}

\noindent\textbf{Vision-Language Models in Remote Sensing. }
While generalist Vision-Language Models (VLMs) like GLM-4.1V~\cite{vteam2025glm45vglm41vthinkingversatilemultimodal} and Qwen3-VL~\cite{Qwen3-VL} exhibit remarkable capabilities, they often struggle with the unique top-down perspectives and scale variations of remote sensing imagery~\cite{luo2024skysensegpt, Zhu_2025_CVPR_skysense_o, zhang2024earthmarker}. Domain-specific models (e.g., GeoChat~\cite{kuckreja2024geochat}, VHM~\cite{pang2025vhm}) address this domain gap but typically operate as black boxes, lacking interpretable reasoning. To introduce transparency, pioneering frameworks like RS-EoT~\cite{shao2025asking_geoeot} and GeoZero~\cite{wang2025geozero} integrate Chain-of-Thought (CoT)~\cite{wei2022chain_cot} for step-by-step geospatial analysis. However, as these models transition from Supervised Fine-Tuning (SFT) to Reinforcement Learning (RL), they predominantly rely on outcome-based rewards. Without explicit step-level verification, they are prone to visual hallucinations, often being rewarded for flawed intermediate logic that coincidentally yields the correct final answer~\cite{liu2025towardsrsthinker, hu2025ringmo, yao2025remotereasoner}. This highlights the critical need for process-supervised alignment in RS.

\noindent\textbf{Process Reward Models. }
Unlike Outcome Reward Models (ORMs)~\cite{cobbe2021training_orm} that evaluate only the final answer, Process Reward Models (PRMs)~\cite{lightman2023let_prm, zheng2025survey_prm} provide granular feedback by assessing each intermediate reasoning step. To bypass the high costs of human annotation seen in early PRMs, recent natural language and math frameworks (e.g., URSA~\cite{luo2025ursa}, GraphPRM~\cite{peng2025rewarding_graphprm}) employ Monte Carlo Tree Search (MCTS)~\cite{browne2012survey_mcts} for automated data synthesis. However, multimodal PRMs remain largely underexplored, particularly within the remote sensing domain. The geospatial field introduces unique challenges where errors frequently arise from subtle visual-textual misalignments rather than purely logical flaws~\cite{pak2025correction_remotesense, lenton2024remotely_remotesense}. GeoPRM bridges this gap by utilizing the Geo-PRM-2M dataset, which is constructed through Automated Process Annotation via Entropy-Guided MCTS and Synthetic Hallucination Injection, ensuring robust token-level verification.

\noindent\textbf{Reinforcement Learning with Tree Search. }
Reinforcement learning, particularly policy optimization algorithms like PPO~\cite{schulman2017proximal_ppo} and GRPO~\cite{deepseekai2025deepseekr1incentivizingreasoningcapability}, has become the standard for aligning VLMs. While integrating process supervision into RL offers clear advantages over outcome supervision~\cite{zheng2025survey_prm, li2025system_prm}, directly using static scalar PRM scores as optimization targets often induces reward hacking and length bias, causing models to generate artificially truncated responses~\cite{luo2025ursa, peng2025rewarding_graphprm}. Concurrently, while tree search algorithms~\cite{ji2025tree_grpo, guo2025segment_spo} are widely used for Test-Time Scaling (TTS), their integration into online RL training is computationally prohibitive. Frameworks like TreeRL~\cite{hou2025treerl} mitigate this via entropy-guided exploration during rollout. Building on these insights, we propose Process-Aware Tree-GRPO. To prevent reward hacking, we replace scalar score accumulations with a drop-moment penalty, and we systematically propagate these robust verification signals through an entropy-guided reasoning tree to optimize the policy efficiently.

\section{Methodology}
\label{sec:methodology}

In this section, we introduce the proposed GeoSolver in detail. We first provide the preliminaries of the base model and Supervised Fine-Tuning (SFT) initialization in Sec.~\ref{sec:preliminaries}. Then, to address the challenge of granular verification, we clarify the formulation of our token-level Process Reward Model (GeoPRM) in Sec.~\ref{sec:prm} and the Process-Aware Tree-GRPO alignment algorithm in Sec.~\ref{sec:tree_grpo}.

\subsection{Preliminaries}
\label{sec:preliminaries}

\noindent\textbf{Base Vision-Language Model.} We build our framework upon the pretrained GLM-4.1V-9B-Base~\cite{vteam2025glm45vglm41vthinkingversatilemultimodal}. This state-of-the-art foundation model features a decoupled architecture ideally suited for geospatial tasks. For visual encoding, it employs Aimv2-Huge~\cite{fini2025multimodal}, a vision transformer capable of handling variable image resolutions and aspect ratios as a critical requirement for processing multi-scale remote sensing imagery. Unlike traditional fixed-resolution encoders, Aimv2-Huge utilizes a dynamic positional encoding scheme that adapts its pretrained position table via bicubic interpolation, allowing it to preserve fine-grained spatial details without distortion. For language modeling, the model incorporates a 3D-RoPE decoder, which further enhances its ability to comprehend complex spatial relationships within the visual context.

\noindent\textbf{Supervised Fine-Tuning.} To equip the base model with the fundamental capability for Chain-of-Thought (CoT) reasoning, we perform supervised fine-tuning on the Geo-CoT380k dataset~\cite{liu2025towardsrsthinker}. This dataset contains 380k samples annotated with structured reasoning paths (e.g., \textit{planning} $\to$ \textit{visual evidence collection} $\to$ \textit{synthesis}). Given an input image $I$, a query $Q$, and the corresponding ground-truth reasoning chain $C = \{c_1, c_2, \dots, c_{|C|}\}$, we train the model to minimize the negative log-likelihood of the target tokens:
\begin{equation}
\mathcal{L}_{\text{SFT}}(\theta) = - \mathbb{E}_{(I, Q, C) \sim \mathcal{D}_{\text{sft}}} \sum_{t=1}^{|C|} \log \pi_\theta(c_t | c_{<t}, I, Q)
\end{equation}
We denote the policy derived from this stage as $\pi_{\text{sft}}$. While $\pi_{\text{sft}}$ successfully adopts the desired reasoning format, it remains susceptible to logical errors and visual hallucinations, serving as the starting policy for our subsequent reinforcement learning alignment.

\subsection{Process Reward Modeling}
\label{sec:prm}

To enable granular verification of reasoning trajectories, we develop GeoPRM, a token-level discriminator designed to assign a scalar likelihood of correctness to each generated token. Unlike step-level reward models, our token-level architecture allows for precise localization of errors such as incorrect bounding box coordinates within the sequence. Based on Geo-CoT380k, we construct Geo-PRM-2M, a composite dataset of approximately 2 million samples, through a dual-strategy pipeline.

\noindent\textbf{Automated Process Annotation via Entropy-Guided MCTS.}
To automate process annotation, recent works in reasoning domains typically rely on Monte Carlo (MC) method~\cite{peng2025rewarding_graphprm, luo2025ursa, luo2024improve_mcts}. However, applying uniform random sampling to the vast multimodal solution space often yields redundant paths that fail to capture the critical divergence points between successful and failed reasoning. To efficiently explore the solution space and establish clear decision boundaries, we design an Entropy-Guided Monte Carlo Tree Search algorithm. This approach prioritizes exploration in regions of high uncertainty within the policy $\pi_{\text{sft}}$.

Tree Construction. We iteratively build the reasoning tree by identifying forking points where the model is uncertain. Specifically, we compute the entropy $H(y_{\text{next}} | I, Q, y_{<t})$ of the next token distribution. We select the top-$N$ tokens with the highest entropy as branching nodes and expand them by performing $T$ rollouts. This process iterates for $K$ rounds, constructing a dense tree of diverse reasoning paths encompassing both correct and flawed logic.

Value Estimation and Labeling. Once the tree is constructed, we estimate the faithfulness of each step $s_t$ based on its rollout success rate. Formally, given the image $I$, query $Q$, and a reasoning prefix $s_{<t}$, the MC value $V(s_t)$ is defined as the proportion of its $T$ rollouts leading to the correct final answer $y_{\text{gt}}$:
    \begin{equation}
        V(s_t) = \frac{1}{T} \sum_{k=1}^T \mathbb{I}(o_k = y_{\text{gt}} | s_{<t}, I, Q)
    \end{equation}
Using this strategy, we generate a pool of 3.72 million reasoning trajectories. To ensure the dataset contributes valid discriminative signals, we calculate the standard deviation of correctness scores within each problem's solution space. We filter out problems with low variance (where the model is either uniformly correct or uniformly incorrect across all paths), retaining 1.37 million high-value samples that effectively highlight the boundary between valid reasoning and plausible errors.

\noindent\textbf{Synthetic Hallucination Injection.}
Apart from logical errors, perceptual inconsistency between images and text is a unique and prominent challenge in multimodal scenarios~\cite{gao2023pal, yan2024errorradar}. While MCTS-based annotation captures logical uncertainty, it may fail to isolate subtle visual-textual misalignments if the model hallucinates its way to the right answer. To explicitly penalize ungrounded visual claims, we propose a targeted synthetic injection engine to construct a subset derived from ground-truth trajectories. We employ two perturbation strategies: 1) Box Perturbation: We manipulate bounding box coordinates to break visual-textual alignment, creating \textit{Small Jitter} (shifting slightly to test precise boundary awareness) and \textit{Large Jitter} (moving to background regions to penalize obvious errors). 2) Fact Modification: We alter factual statements, such as object counts or attributes, creating \textit{Tampered} negatives that violate the visual premise.

Combined with unaltered positive anchors from Geo-CoT380k, this synthetic engine contributes approximately 0.7 million samples, bringing the total Geo-PRM-2M training instances to 2 million.

\noindent\textbf{Training Objective.}
We initialize GeoPRM from $\pi_{\text{sft}}$ and append a linear binary classification head. To evaluate the current token within the full reasoning context, we concatenate the image $I$ and query $Q$ with all preceding tokens. Formally, we define the training as a dense token-level binary classification task over a sequence of length $L$:
\begin{equation}
    \min_\theta \mathcal{L}(\theta) = - \frac{1}{L} \sum_{t=1}^{L} m_t \left[ y_t \log \hat{y}_t + (1 - y_t) \log (1 - \hat{y}_t) \right]
\end{equation}
where $y_t$ is the label for the $t$-th token, $\hat{y}_t = P(y_t | I, Q, y_{<t})$ is the predicted probability, and the mask $m_t$ restricts the loss calculation to valid reasoning tokens.

\begin{figure}[tb]
    \centering
    \includegraphics[width=0.95\linewidth]{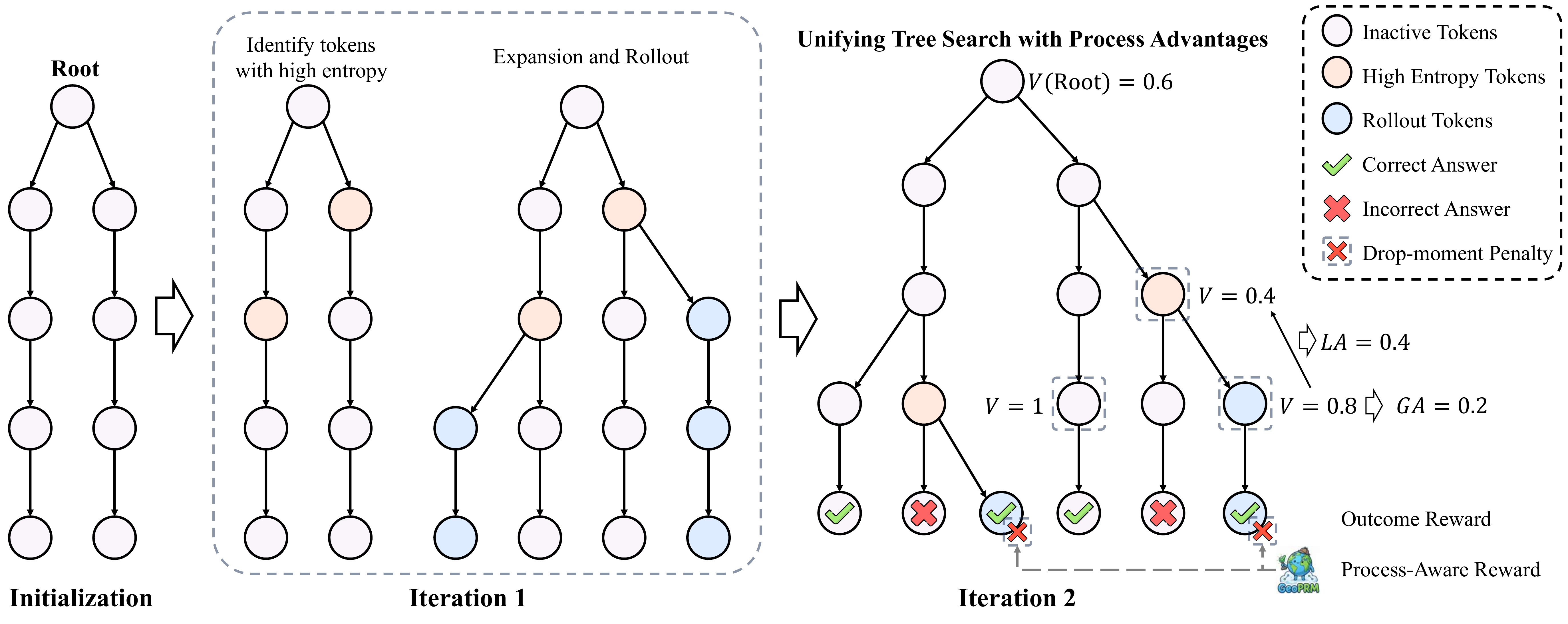}
    \vspace{-0.3cm}
    \caption{Illustration of the Process-Aware Tree-GRPO. The reasoning tree dynamically expands by identifying high-entropy tokens, followed by trajectory rollouts. Leaf nodes receive outcome rewards which are further refined by GeoPRM's drop-moment penalty. These process-aware signals are then aggregated upwards to compute Local Advantage ($LA$) and Global Advantage ($GA$) for fine-grained policy optimization.}
    \label{fig:tree_method}
    \vspace{-0.5cm}
\end{figure}

\subsection{Process-Aware Tree-GRPO}
\label{sec:tree_grpo}

Vanilla GRPO normalizes in-group rewards based solely on final outcomes. This presents two critical limitations: 1) Reward Sparsity and Hacking: Outcome rewards ignore intermediate quality. Conversely, directly using scalar PRM scores as optimization targets induces a length bias. Because PRMs are trained on distributions where longer reasoning chains inherently carry a higher statistical risk of accumulating errors, they naturally tend to assign lower scores to longer sequences. 2) Inefficient Exploration: Independent linear sampling fails to leverage the hierarchical and compositional nature of spatial reasoning. 
To address these limitations, we present Process-Aware Tree-GRPO, integrating entropy-guided tree search with a drop-moment penalty mechanism, as illustrated in Figure~\ref{fig:tree_method}.

\noindent\textbf{Process-Aware Reward via Drop-Moments.}
To extract reliable verification signals from GeoPRM, we adopt the concept of the drop-moment~\cite{luo2025ursa}, signifying a sudden loss of confidence between consecutive steps. For a trajectory $o_i$ with a GeoPRM reward sequence $\{r_1^i, r_2^i, \dots, r_{L_i}^i\}$, a drop-moment occurs if:
\begin{equation}
    \delta^i_p = \max_{j} \left( r_{j-1}^i - r_j^i \right) > \rho
\end{equation}
where $\rho$ is the sensitivity threshold. In the remote sensing domain, the final outcome of a trajectory is typically evaluated by a continuous domain-specific scoring mechanism, yielding an outcome score $S_{\text{out}}(o_i) \in [0, 1]$ (e.g., IoU or mAP). When a drop-moment is detected, we apply a penalty factor $\gamma \in (0, 1)$ to this base score:
\begin{equation}
    \label{eq:reward}
    R = 
    \begin{cases} 
    S_{\text{out}}(o_i), & \text{if } \delta^i_p < \rho \\
    \gamma \cdot S_{\text{out}}(o_i), & \text{if } \delta^i_p \ge \rho
    \end{cases}
\end{equation}
This formulation circumvents the inherent length bias by focusing on relative confidence drops rather than absolute score accumulations, ensuring the policy learns from trajectories that are both outcome-accurate and process-rigorous.

\noindent\textbf{Unifying Tree Search with Process Advantages.}
Although the drop-moment mechanism effectively refines the trajectory reward $R(o_i)$ based on process quality, this scalar remains a sequence-level reward. To explicitly assign credit to intermediate reasoning tokens, we must distribute this reward along the reasoning path. Therefore, during the rollout stage, we employ the aforementioned entropy-guided tree construction to yield a hierarchical reasoning tree $\mathcal{T}$ for a problem (see the left of Figure~\ref{fig:tree_method}). 

For any intermediate step $s_n$, let $L(s_n)$ denote the set of all complete leaf trajectories descending from it. The value of step $s_n$ is computed as the average process-aware reward of its descendants:
\begin{equation}
    V(s_n) = \frac{1}{|L(s_n)|} \sum_{l \in L(s_n)} R(l)
\end{equation}
By aggregating the drop-moment penalized rewards from the leaves, $V(s_n)$ accurately reflects the potential of step $s_n$ to lead to a faithful and correct conclusion. Based on this value, we define the Global Advantage as $GA(s_n) = V(s_n) - V(\text{root})$, measuring the node's performance relative to the overall tree expectation; and the Local Advantage as $LA(s_n) = V(s_n) - V(p(s_n))$, measuring the improvement over its immediate parent $p(s_n)$. The final re-weighted advantage is:
\begin{equation}
    \hat{A}(s_n) = \frac{GA(s_n) + LA(s_n)}{\sqrt{|L(s_n)|}}
\end{equation}
Because non-leaf steps appear in multiple complete sequences and would be repeatedly computed during optimization, we downweight their advantage by dividing by $\sqrt{|L(s_n)|}$, the square root of the number of leaf nodes in the subtree. This critical adjustment prevents the policy from overfitting to early, shared reasoning stages.

Finally, we update the policy $\pi_\theta$ using the clipped surrogate objective over all nodes in the tree:
\begin{equation}
\begin{split}
    \mathcal{L}_{\text{Tree-GRPO}}(\theta) &= -\mathbb{E}_{(I, Q)\sim\mathcal{D}, \mathcal{T} \sim \pi_{\theta_{old}}} \\
    \frac{1}{|\mathcal{T}|} \sum_{s_n \in \mathcal{T}} &\min \Bigg( r_\theta(s_n) \hat{A}(s_n), \text{clip}\left(r_\theta(s_n), 1-\epsilon, 1+\epsilon\right) \hat{A}(s_n) \Bigg) 
\end{split}
\end{equation}
where $r_\theta(s_n) = \frac{\pi_\theta(s_n)}{\pi_{\theta_{old}}(s_n)}$ denotes the probability ratio. By explicitly linking drop-moments to leaf rewards and propagating them via local and global advantages through the tree, this algorithm systematically aligns the model toward verifiable, hallucination-free reasoning.

\section{Experiments}
\label{sec:experiments}

\subsection{Experimental Setup}
\label{sec:setup}

\begin{table}[!tb]
\begin{center}
\resizebox{\linewidth}{!}{
\begin{tabular}{lcccccccccccc}
\toprule
\multirow{2}{*}{\textbf{Method}} 
& \multicolumn{4}{c}{\textbf{Visual Grounding}}
& \multicolumn{2}{c}{\textbf{Object Detection}} 
& \multicolumn{4}{c}{\textbf{Object Counting}} 
\\
\cmidrule(lr){2-5} \cmidrule(lr){6-7} \cmidrule(lr){8-11}
 & \textbf{DIOR} &  \textbf{RRSIS} &  \textbf{VRS} & \textbf{RSVG} & \textbf{DOTA} & \textbf{HRRSD} & \textbf{RSOD} & \textbf{VHR} & \textbf{DOTA} & \textbf{HRRSD} \\
\midrule
\multicolumn{6}{l}{\textcolor{gray}{\textit{Closed-source Commercial Vision-Language Models}}} \\
Claude-sonnet-4~\cite{anthropic2025claude4_opus_sonnet} & 8.87 & 11.62 & 6.65 & 2.53 & 3.89  & 14.87  &  51.5 & 25.0 & 25.17 & 50.11  \\
Gemini-2.0-flash~\cite{comanici2025gemini} & 11.40 & 16.05 & 13.52 & 1.75 & 14.30  & 28.92  & \textbf{63.5} & 39.0 & 29.36 & 54.65    \\
ChatGPT-5~\cite{openai2025gpt5_blog}  & 8.27 & 10.60 & 5.65 & 1.49 & 8.66  & 13.15  &  40.0 & 58.0 & \underline{36.20} & 58.50   \\

\midrule
\multicolumn{6}{l}{\textcolor{gray}{\textit{Open-source Reasoning Vision-Language Models}}} \\
Kimi-VL-Thinking~\cite{kimiteam2025kimivltechnicalreport} & $-$ & $-$  & $-$ & $-$ & $-$ & $-$ &  15.5 & 53.0 & 30.68 & 46.26   \\
GLM-4.1V-Thinking~\cite{vteam2025glm45vglm41vthinkingversatilemultimodal} & 39.41 & 43.03 & \underline{43.20} & \underline{19.35} & \underline{40.45} & \underline{55.53}  &  28.5 & \underline{62.5} & 29.80 & 58.96  \\
RS-EoT~\cite{shao2025asking_geoeot} & 9.23 & 12.09 & 1.28 & 11.79 & $-$ & $-$ & 35.0 & 49.5 & 28.04 & 57.62  \\
\midrule 
\multicolumn{6}{l}{\textcolor{gray}{\textit{Open-source Remote Sensing Vision-Language Models}}} \\
GeoChat~\cite{kuckreja2024geochat} &  29.87 & 30.30 & 23.66 & 0.13 &  $-$ & $-$ & 23.5 & 47.0 & 29.87 & 43.22   \\
VHM~\cite{pang2025vhm}  & \underline{49.90} & \underline{55.20} & 34.91 & 5.80 & 2.37  & 12.47  &  16.0 & 48.5 & 32.67 & 46.71    \\
SkySenseGPT~\cite{luo2024skysensegpt} & 27.76 & 33.63 & 21.25 & 4.57 & 4.56  & 6.23  &  \underline{51.5} & 49.5 & 33.11 & 58.73  \\
EarthDial~\cite{soni2025earthdial}  &  39.46 & 35.39 & 13.04 & 7.18 & 3.52 & 8.05 & 41.0 & 52.5 & 32.23 & \underline{61.45}  \\
\midrule

\textbf{GeoSolver} & \textbf{75.62} & \textbf{76.66} & \textbf{64.57} & \textbf{27.43} & \textbf{53.47} & \textbf{94.74} &  45.5 & \textbf{79.0} & \textbf{45.92} & \textbf{84.13}  \\
\bottomrule
\end{tabular}
}
\caption{Comparison of GeoSolver with state-of-the-art models on object-level remote sensing tasks. We report mIoU and mAP@50 for Visual Grounding (DIOR-RSVG\nocite{zhan2023rsvg}, RRSIS-D\nocite{liu2024rotated}, VRSBench\nocite{NEURIPS2024_05b7f821}, and RSVG\nocite{sun2022visual}) and Object Detection (DOTAv2-val\nocite{xia2018dota, Ding_2019_CVPR, 9560031} and HRRSD\nocite{zhang2019hierarchical}), alongside Accuracy for Object Counting (RSOD\nocite{long2017accurate}, NWPU-VHR\nocite{cheng2014multi}, DOTAv2-val\nocite{xia2018dota, Ding_2019_CVPR, 9560031}, and HRRSD\nocite{zhang2019hierarchical}). The best results are bolded, and the second-best are underlined.}
\label{tab: exp_main_obj}
\end{center}
\vspace{-1.2cm}
\end{table}

\noindent\textbf{Datasets and Tasks.}
To comprehensively validate the versatility of GeoSolver, we evaluate its performance across six major remote sensing tasks: Object Counting (OC), Object Detection (OD), Visual Grounding (VG), Scene Classification (SC), Visual Question Answering(VQA) and Image Captioning(IC). These tasks are assessed on a diverse suite of 17 established benchmark datasets. For training, we utilize the Geo-CoT380k dataset~\cite{liu2025towardsrsthinker} for the initial SFT stage, and its expanded training set for the Tree-GRPO alignment. A detailed breakdown of all datasets, splits, and metrics is provided in the supplementary material.

\noindent\textbf{Baselines.}
We benchmark GeoSolver against a broad spectrum of state-of-the-art models, categorized into three groups: (1) leading closed-source commercial systems (e.g., Gemini-2.0-Flash~\cite{comanici2025gemini}, Claude-3.5-Sonnet~\cite{anthropic2025claude4_opus_sonnet}); (2) open-weight general and domain-specific VLMs (e.g., Qwen2.5-VL~\cite{Qwen2.5-VL}, GeoChat~\cite{kuckreja2024geochat}, EarthDial~\cite{soni2025earthdial}); and (3) recent reasoning-centric or CoT-enabled frameworks (e.g., GLM-4.1V-9B-Thinking~\cite{vteam2025glm45vglm41vthinkingversatilemultimodal}, RS-EoT-7B~\cite{shao2025asking_geoeot}). Furthermore, for Test-Time Scaling (TTS) evaluations, we compare GeoPRM against existing general mathematical PRMs and general VLMs acting as verifiers. Comprehensive descriptions of these baselines can be found in the supplementary material.

\noindent\textbf{Implementation Details.}
GeoSolver is initialized from the GLM-4.1V-9B-Base~\cite{vteam2025glm45vglm41vthinkingversatilemultimodal} checkpoint. All training stages are conducted on a cluster of four NVIDIA H200 GPUs. Specifically, GeoPRM is trained on our Geo-PRM-2M dataset for 2 epochs with a batch size of 128. The policy model undergoes 1 epoch of SFT, followed by 1000 optimization steps of Process-Aware Tree-GRPO. Full training hyperparameters are detailed in the supplementary material.

\begin{table}[!t]
\begin{center}
\resizebox{0.98\linewidth}{!}{
\begin{tabular}{lcccccccccccc}
\toprule
\multirow{2}{*}{\textbf{Method}} 
& \multicolumn{5}{c}{\textbf{Scene Classification}}
& \multicolumn{2}{c}{\textbf{VQA}} 
& \multicolumn{3}{c}{\textbf{Image Caption}} 
\\
\cmidrule(lr){2-6} \cmidrule(lr){7-8} \cmidrule(lr){9-11}
 & \textbf{NWPU} &  \textbf{AID} &  \textbf{RS19} & \textbf{SIRI} & \textbf{UCM} & \textbf{VRS} & \textbf{HR} & \textbf{RSIT} & \textbf{NWPU} & \textbf{RSIC} \\
\midrule
\multicolumn{6}{l}{\textcolor{gray}{\textit{Closed-source Commercial Vision-Language Models}}} \\
Claude-sonnet-4 & 58.44 & 60.33 & 76.32 & 64.33 & 67.86 & 58.95 & 55.94 & 20.14 & 28.32 & 11.58  \\
Gemini-2.0-flash & 74.89 & 76.00 & 90.00 & 72.00 & 85.95 & 61.92 & 49.95 & 15.73 & 20.55 & 10.85    \\
ChatGPT-5  & 82.22 & 75.50 & \underline{95.53} & 75.00 & 88.57 & 62.20 & 65.94 & 27.27 & 39.62 & 16.83  \\
\midrule
\multicolumn{6}{l}{\textcolor{gray}{\textit{Open-source Vision-Language Models}}} \\
MiniGPT-v2 & 32.67 & 27.17 & 30.79 & 26.67 & 32.86 & 37.35 & 50.95 & 25.45 & 37.75 & 15.40 \\
Qwen2.5-VL & 68.89 & 71.67 & 86.05 & 67.33 & 78.33 & 59.64 & 57.43 & 27.92 & 38.89 & 17.80  \\
\midrule
\multicolumn{6}{l}{\textcolor{gray}{\textit{Open-source Reasoning Vision-Language Models}}} \\
Kimi-VL-Thinking & 72.22 & 70.50 & 88.68 & 69.00 & 77.62 & \underline{69.09} & 70.93 & 24.82 & 34.84 & 15.60 \\
GLM-4.1V-Thinking & 70.09& 69.67 & 86.84 & 60.33 & 82.86 & 63.57 & 55.94 & 20.57 & 29.59 & 12.57   \\
RS-EoT & 75.56 & 75.83 & 90.25 & \textbf{78.88} & 85.95 & 69.00 & 70.05 & 25.16 & 37.27 & 23.02 \\
\midrule 
\multicolumn{6}{l}{\textcolor{gray}{\textit{Open-source Remote Sensing Vision-Language Models}}} \\
VHM  & \underline{91.33} & \underline{79.00} & 91.84 & 64.33 & \underline{89.29} & 54.26 & 69.43 & 38.93 & 50.69 & 25.66   \\
SkySenseGPT &83.33 &75.50 & 93.16 & 55.33 & 85.00 & 48.59 & 63.44 & 37.76 & 23.33 & \textbf{42.47}  \\
EarthDial  & 76.67 & 67.33 & 88.76 & 73.42 & 80.71 & 39.18 & \underline{72.43} & \underline{42.09} & \underline{67.14} & 29.09 \\

\midrule

\textbf{GeoSolver} & \textbf{96.44} & \textbf{98.33} & \textbf{99.50} & \underline{76.00} & \textbf{92.38} & \textbf{77.02} & \textbf{73.19} & \textbf{52.50} & \textbf{80.93} & \underline{36.18} \\
\bottomrule
\end{tabular}
}
\caption{Comparison of GeoSolver with state-of-the-art models on scene-level tasks. We report Accuracy for Scene Classification (NWPU-RESISC45\nocite{cheng2017remote}, AID\nocite{xia2017aid}, WHU-RS19\nocite{xia2010structural}, SIRI-WHU\nocite{zhao2016dirichlet, zhao2016fisher, zhu2016bag}, and UCMerced\nocite{Nilsback08}) and VQA (VRSBench\nocite{NEURIPS2024_05b7f821} and RSVQA-HR\nocite{lobry2020rsvqa}). For Image Captioning (RSITMD\nocite{yuan2021exploring}, NWPU-Captions\nocite{cheng2022nwpu}, and RSICD\nocite{lu2017exploring}), we report the BLEU-4 metric. }
\label{tab: exp_main_scene}
\end{center}
\vspace{-1.2cm}
\end{table}

\subsection{Comparison with State-of-the-Art}

We first evaluate the standard inference performance of GeoSolver. Table~\ref{tab: exp_main_obj} and Table~\ref{tab: exp_main_scene} present the quantitative results across object-level and scene-level tasks, respectively. Despite the absence of inference-time search, GeoSolver consistently outperforms existing domain-specific models like GeoChat~\cite{kuckreja2024geochat} and VHM~\cite{pang2025vhm}. Notably, on fine-grained tasks such as Visual Grounding and Object Detection, GeoSolver achieves significant performance margins over general-purpose reasoning models (e.g., GLM-4.1V-Thinking~\cite{vteam2025glm45vglm41vthinkingversatilemultimodal}). This demonstrates that the Process-Aware Tree-GRPO training endows the policy with a strong intrinsic capability for faithful spatial reasoning, effectively curbing object hallucinations during standard autoregressive generation.

\subsection{Scaling Test-Time Reasoning with GeoPRM}
\label{sec:tts}

To investigate the potential of compute-optimal inference in remote sensing, we comprehensively evaluate Test-Time Scaling (TTS) enabled by our process reward model, GeoPRM. We design three sets of experiments to assess the efficacy of different search strategies~\cite{snell2024scaling_beam}, the scaling behavior with increased compute budgets, and the superiority of our domain-specific PRM over generic verifiers. Note that for all aggregated task columns in this section, the reported values represent the average performance across their corresponding datasets. Detailed per-dataset metrics and scaling curves for additional tasks can be found in the supplementary material.

\begin{figure}[!tb]
    \centering
    
    \begin{minipage}[c]{0.63\textwidth}
        \makeatletter\def\@captype{table}\makeatother 
        \centering

        \resizebox{\linewidth}{!}{
        \begin{tabular}{lcccccccc}
        \toprule
        \textbf{Method} &  \textbf{VG } &  \textbf{Detect} & \textbf{OC}  &  \textbf{ SC } & \textbf{VQA} &  \textbf{IC} &  \(\uparrow\) (Avg)  \\
        \midrule
        GeoSolver (w/o TTS) & 58.19 & 74.11 & 54.06 & 90.62 & 70.07 & 47.00 & $-$ \\
        \midrule
        Self-Consistency & 59.44 & 75.17 & 59.28 & 92.57 & 76.58 & 47.51 &  4.1\% \\
        Best-of-N & 66.35 & 82.51 & 73.92 & 96.01 & \textbf{88.70} & \textbf{48.18} &  15.8\% \\
        Beam Search & \textbf{68.04} & \textbf{84.66} & \textbf{75.45} & \textbf{98.39} & 87.84 & 48.05 & \textbf{17.5\%} \\
        \bottomrule
        \end{tabular}
        }
        \vspace{-0.3cm}
        \caption{Comparison of different verification strategies on GeoSolver. We evaluate greedy decoding (GeoSolver w/o TTS), Self-Consistency (majority voting), and our GeoPRM utilizing both Best-of-N and Beam Search strategies. The generation budget is set to 32.}
        \label{tab: exp_tts_geosolver}

        \hfill
        
        \resizebox{\linewidth}{!}{%
        \begin{tabular}{lccccccc}
        \toprule
        \textbf{Method} & \textbf{VG} &  \textbf{Detect} &  \textbf{OC} & \textbf{SC} & \textbf{VQA} & \textbf{IC}  & \textbf{Avg}  \\
        \midrule
        Self-Consistency &  \underline{59.44} & 75.17 & \underline{59.28} & \underline{92.57} & \underline{76.58} & 47.51 & \underline{68.43}  \\
        \midrule
        \multicolumn{8}{l}{\textcolor{gray}{\textit{Open-source Vision-Language Models}}} \\
        Qwen3-VL-8B & 54.20 & \underline{75.25} & 50.23 & 91.09 & 70.45 & \underline{47.72} & 64.82 \\
        GLM-4.1V & 53.01 & 74.62 & 47.36 & 90.78 & 71.85 & 47.24 & 64.14 \\
        \midrule
        \multicolumn{8}{l}{\textcolor{gray}{\textit{Open-source Reward Models}}} \\
        GraphPRM & 53.82 & 72.92 & 42.36 & 89.58 & 68.35 & 47.17 & 62.37 \\
        URSA-8B-RM & 53.27 & 73.30 & 45.15 & 86.43 & 70.50 & 46.98 & 62.61 \\
        \midrule
        GeoPRM & \textbf{66.35} & \textbf{82.51} & \textbf{73.92} & \textbf{96.01} & \textbf{88.70} & \textbf{48.18} & \textbf{75.95}   \\
        \bottomrule

        \end{tabular}%

        }
        \vspace{-0.3cm}
        \caption{Cross-model reward model comparison using BoN performance ($N=32$). We evaluate the verification efficacy of various strategies, including Self-Consistency, generic VLMs acting as verifiers, open-source PRMs, and our domain-specific GeoPRM across diverse remote sensing tasks. }
        \label{tab: exp_tts_all}
        
    \end{minipage}\hfill 
    \begin{minipage}[c]{0.33\textwidth}
        \makeatletter\def\@captype{figure}\makeatother 
        \centering
        
        \begin{subfigure}{\linewidth}
        \centering
        \includegraphics[width=0.95\linewidth]{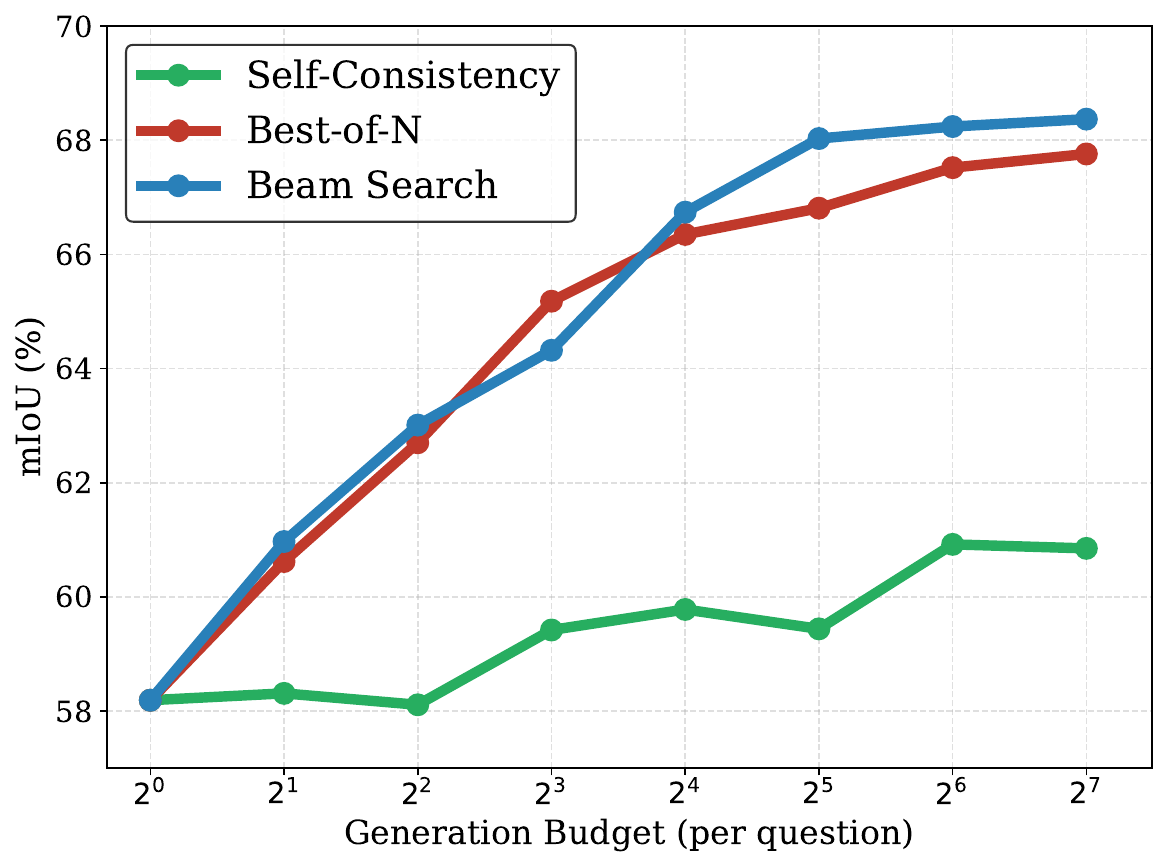}
        \caption{Visual Grounding (VG)}
        \label{fig:tts_geosolver_result_a}
        \end{subfigure}
        
        \vspace{0.2cm} 
        
        \begin{subfigure}{\linewidth}
            \centering
            \includegraphics[width=0.95\linewidth]{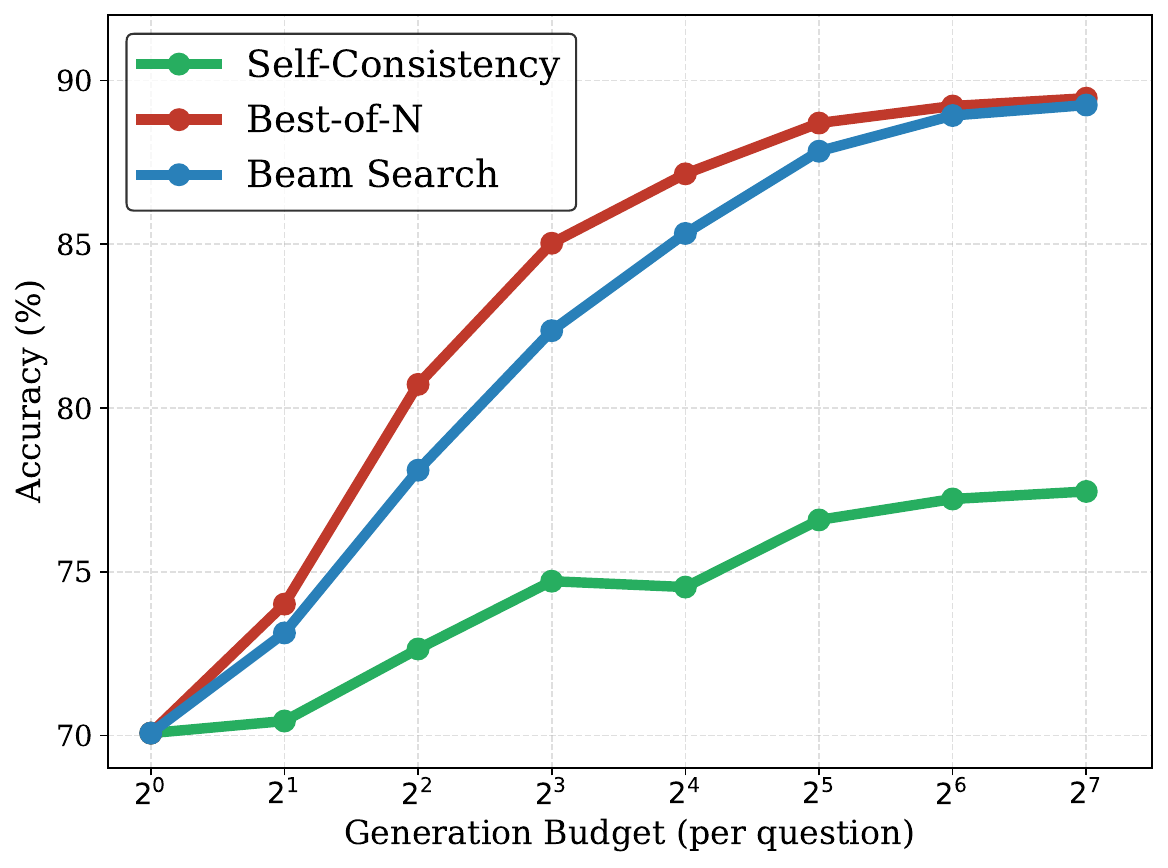}
            \caption{VQA}
            \label{fig:tts_geosolver_result_b}
        \end{subfigure}
        
        \caption{Compute-optimal scaling behavior of GeoSolver equipped with GeoPRM. The curves illustrate the performance gains on (a) VG and (b) VQA as the compute budget ($N$) increases.}
        \label{fig:tts_geosolver_results}
    \end{minipage}
    \vspace{-0.8cm}
\end{figure}

\noindent\textbf{Efficacy of Verification Strategies.} 
We first evaluate the impact of different test-time verification methods applied directly to our GeoSolver policy. As shown in Table~\ref{tab: exp_tts_geosolver}, relying solely on greedy decoding (w/o TTS) yields a solid baseline, but lacks a mechanism to correct intermediate logical flaws. While Self-Consistency (majority voting) provides marginal improvements, it is fundamentally bottlenecked by the base policy's distribution. In contrast, integrating GeoPRM as the verifier, unlocking explicit process supervision, leads to substantial gains. Both Best-of-N and Beam Search strategies consistently elevate performance across all metrics, with Beam Search demonstrating particular efficacy in densely grounded tasks like Object Counting and Detection by pruning hallucinatory branches early in the reasoning tree.

\noindent\textbf{Compute-Optimal Scaling Behavior.} 
A hallmark of effective test-time reasoning is predictable performance scaling concerning the inference compute budget. We vary the generation budget to observe this scaling law. As illustrated in Figure~\ref{fig:tts_geosolver_results}, scaling the test-time budget consistently and monotonically improves performance. Notably, on complex reasoning tasks such as VG and VQA, the performance curves exhibit a robust log-linear improvement before plateauing, confirming that GeoPRM effectively harnesses additional compute to navigate complex geospatial logic without suffering from reward over-optimization.

\noindent\textbf{Cross-Model Reward Model Comparison.} 
A critical determinant of TTS efficacy is the domain alignment of the reward model itself. In Table~\ref{tab: exp_tts_all}, we benchmark GeoPRM against alternative verification systems, including general-purpose VLMs acting as verifiers and state-of-the-art out-of-domain mathematical PRMs like URSA-8B-RM~\cite{luo2025ursa}. The results reveal a stark contrast: general-domain PRMs and standard VLMs struggle to accurately assess geospatial coordinate logic and dense visual relations, often underperforming simple Self-Consistency. Conversely, GeoPRM provides highly calibrated, token-level feedback for spatial relations and visual evidence, empirically validating the absolute necessity of our domain-specific Geo-PRM-2M construction pipeline.

\begin{figure}[!tb]
    \centering
    \begin{subfigure}{0.48\textwidth}
        \centering
        \includegraphics[width=\linewidth]{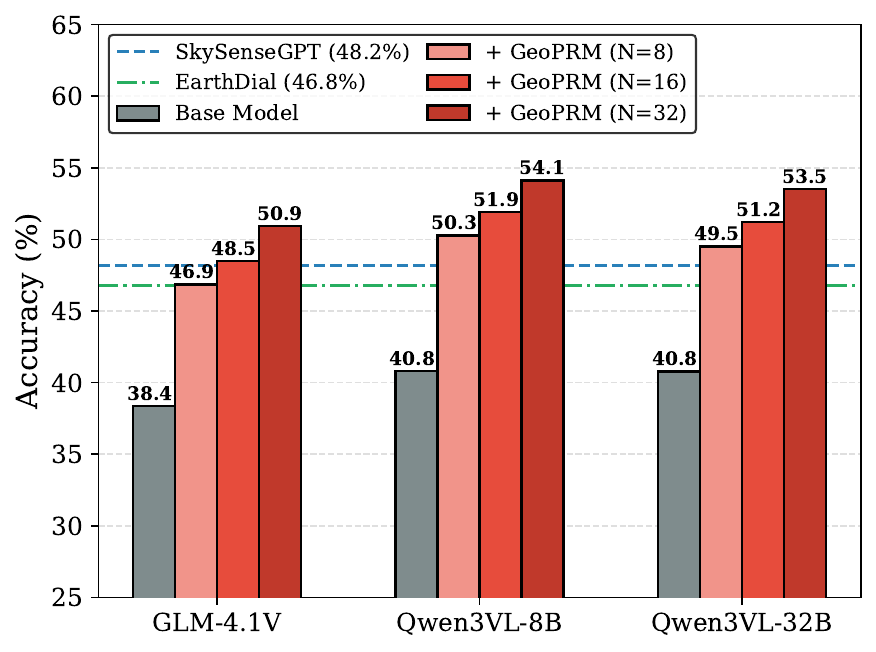} 
        \caption{Object Counting (OC)}
        \label{fig:tts_general_result_a}
    \end{subfigure}
    \hfill 
    \begin{subfigure}{0.48\textwidth}
        \centering
        \includegraphics[width=\linewidth]{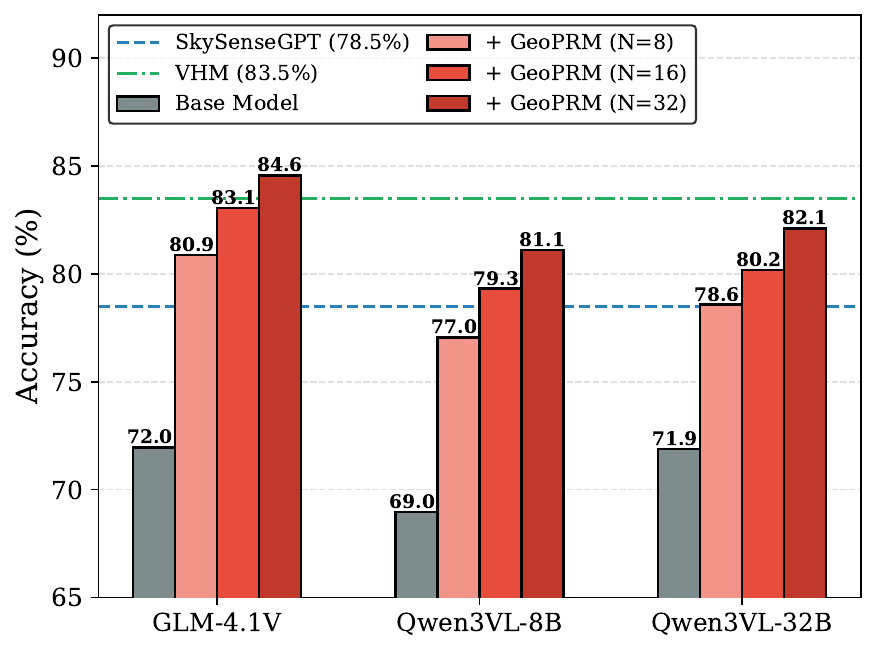} 
        \caption{Scene Classification (SC)}
        \label{fig:tts_general_result_b}
    \end{subfigure}
    \vspace{-0.2cm}
    \caption{Cross-model generalization capabilities of GeoPRM using BoN performance. The gray bars indicate the base model performance, while the red gradient bars show consistent accuracy gains as the generation budget ($N$) increases. Horizontal dashed lines represent the baseline performance of fully fine-tuned remote sensing models. Notably, with $N=32$, GeoPRM-guided generalists surpass these domain experts.}
    \label{fig:tts_general_results}
    \vspace{-0.2cm}
\end{figure}

\subsection{Cross-Model Generalization}
\label{sec:cross_model}

A compelling indicator of a robust Process Reward Model is its ability to generalize beyond the specific policy it was trained alongside. In this section, we investigate whether GeoPRM can function as a universal verifier by scoring the inference rollouts of general-purpose VLMs. Specifically, we apply GeoPRM to guide GLM-4.1V~\cite{vteam2025glm45vglm41vthinkingversatilemultimodal}, Qwen3-VL-8B, and Qwen3-VL-32B~\cite{Qwen3-VL}. Remarkably, as illustrated in Figure~\ref{fig:tts_general_results}, equipping these generalist base models with GeoPRM via Test-Time Scaling ($N \in \{8, 16, 32\}$) yields substantial and consistent performance improvements over their greedy decoding. More importantly, with a sufficient compute budget (e.g., $N=32$), these general-purpose models successfully surpass the performance of fully fine-tuned, domain-specific remote sensing models (such as SkySenseGPT~\cite{luo2024skysensegpt}, EarthDial~\cite{soni2025earthdial}, and VHM~\cite{pang2025vhm}), which are denoted by the horizontal reference lines. This striking phenomenon underscores that GeoPRM has not merely overfit to GeoSolver's output distribution; rather, it has successfully internalized a generalized, transferable discriminative logic for multimodal geospatial verification.

\begin{table}[!tb]
\begin{center}
\resizebox{\linewidth}{!}{
\begin{tabular}{lccc|cccccc|c}    
\toprule
\multirow{2}{*}{\textbf{Models}} & \multirow{2}{*}{\textbf{Tree}} &  \multicolumn{2}{c|}{\textbf{PRM}}  &     \textbf{VG}  & \textbf{OC} & \textbf{Det} & \textbf{IC}  &  \textbf{SC} & \textbf{VQA} & \multirow{2}{*}{\textbf{ \space Avg \space }}  \\
\cmidrule(lr){3-4}
 &  & \textbf{APS}  & \textbf{PA} & (mIoU) & (Acc) & (mAP) & (BLEU-4) & (Acc) & (Acc) \\
\midrule

Base (GLM-4.1V-Base) & $-$ & $-$ & $-$ & 34.48 & 16.96 & 3.56 & 12.43 & 59.78 & 8.16 & 22.56 \\

 + SFT & $-$ & $-$ & $-$  & 53.77 & 49.73 & 59.42 & 44.90 & 86.01 & 68.23 & 60.34 \\

\midrule
Vanilla GRPO & \xmark & \xmark & \xmark & 59.31 & 55.56 & 68.89 & 54.89 & 91.74 & 72.17 & 67.09 \\
 + Average Process Score   &  \xmark & \checkmark & \xmark & 48.69 & 48.22 & 50.76 & 46.45 & 87.21 & 67.44 & 58.13 \\
 + Process-Aware   &  \xmark & \xmark & \checkmark & \underline{60.12} & \underline{59.21} & 69.90 & 54.67 & \underline{92.01} & 73.02 & \underline{68.16}   \\
 + Tree-based   &  \checkmark & \xmark & \xmark & 59.46 & 57.77 & \underline{70.82} & \underline{55.21} & 91.83 & \underline{73.68} & 68.13    \\
Process-Aware Tree-GRPO &  \checkmark & \xmark & \checkmark & \textbf{61.07} & \textbf{63.6} & \textbf{73.97} & \textbf{56.54} & \textbf{92.73} & \textbf{75.11} & \textbf{70.51} \\

\bottomrule
\end{tabular}
}
\caption{Ablation study of different RL alignment strategies on GeoSolver. We evaluate the progressive integration of Vanilla GRPO, Average Process Score (APS), our Process-Aware (PA), and Tree-based exploration. }
\label{tab: exp_abl_model}
\end{center}
\vspace{-0.9cm}
\end{table}

\subsection{Ablation Study}
\label{sec:ablation}

\begin{table}[!tb]
\begin{center}
\resizebox{0.9\linewidth}{!}{
\begin{tabular}{llcccccccc}    
\toprule
\multirow{2}{*}{\textbf{Models}} & \multirow{2}{*}{\textbf{Dataset}} &  \multicolumn{4}{c}{\textbf{Object-level}} & \multicolumn{4}{c}{\textbf{Scene-level}}   \\
\cmidrule(lr){3-6} \cmidrule(lr){7-10}
 &  &  N$=$4 &  N$=$8 &  N$=$16 & N$=$32 &  N$=$4 &  N$=$8 &  N$=$16 & N$=$32 \\
\midrule

\multirow{3}{*}{GeoSolver} & Geo-PRM-2M & \textbf{68.44} & \textbf{70.92} & \textbf{73.26} & \textbf{74.26} & 74.03 & \textbf{75.87} & \textbf{76.97} & \textbf{77.63} \\
  &  w/o $S_{MC}$ & 62.74 & 63.60 & 63.82 & 64.27 & 69.53  & 70.21  & 69.82  &  70.77  \\
  &  w/o $S_{SHI}$ & 66.28 & 67.81 & 69.72 & 70.26 &  \textbf{74.24} & 75.46 & 76.37 & 76.89  \\
\midrule
\multirow{3}{*}{Qwen3-VL-8B} & Geo-PRM-2M & \textbf{28.70} & \textbf{30.98} & \textbf{32.56} & \textbf{34.13} & \textbf{48.51} & \textbf{50.45} & \textbf{51.93} & \textbf{52.99} \\
  &  w/o $S_{MC}$ & 24.87 & 25.38  & 25.22 & 26.11 & 45.11 & 45.88 & 46.04 & 45.97   \\
  &  w/o $S_{SHI}$ & 26.85 & 27.88 & 28.30 & 28.92 & 47.80 & 49.57 & 50.24 & 50.78 \\

\bottomrule
\end{tabular}
}
\caption{Ablation of Geo-PRM-2M training data components evaluated using BoN performance. We compare reward models trained on the full dataset against those trained without intrinsic MCTS samples (w/o $S_{MC}$) or without Synthetic Hallucination Injection (w/o $S_{SHI}$). Performance is evaluated on both GeoSolver and generic model (Qwen3-VL-8B) across scaling budgets ($N$).}
\label{tab: exp_abl_dataset}
\end{center}
\vspace{-1.2cm}
\end{table}

To systematically validate our proposed components, we ablate the RL alignment strategies and the PRM training data composition.

\noindent\textbf{Reward Formulation and Tree Exploration.} 
In Table~\ref{tab: exp_abl_model}, we systematically ablate the components of our RL alignment phase. First, compared to Vanilla GRPO (only outcome rewards), integrating a naive Average Process Score (APS) induces severe reward hacking: the policy artificially truncates reasoning chains to minimize accumulated penalties. Conversely, our Process-Aware (PA) drop-moment mechanism circumvents this length bias by penalizing only sudden relative confidence drops, effectively filtering logical errors while preserving reasoning completeness. 
Independent of the reward formulation, we investigate the impact of the exploration structure. Upgrading Vanilla GRPO with our entropy-guided tree search yields substantial gains over chain-based rollouts. Under an identical token budget, the tree structure enables a much denser exploration of critical branching points. Ultimately, synergizing these two structural and formulation improvements into the unified Process-Aware Tree-GRPO pipeline achieves the highest overall performance.

\noindent\textbf{Efficacy of PRM Synthesis Strategies.} 
A robust PRM requires both rigorous logical evaluation and precise visual grounding. In Table~\ref{tab: exp_abl_dataset}, we ablate the two core synthesis strategies of Geo-PRM-2M: intrinsic MCTS samples ($S_{MC}$) and Synthetic Hallucination Injection ($S_{SHI}$). Training the reward model without $S_{SHI}$ diminishes its sensitivity to subtle visual-textual misalignments, confirming that explicit box perturbations are vital for detecting ungrounded coordinates. Similarly, omitting $S_{MC}$ severely limits the model's capacity to evaluate deep, multi-step logical reasoning. The consistent TTS performance drop across both GeoSolver and the generic Qwen3-VL-8B confirms that intrinsic logical mining and synthetic visual perturbations are mutually indispensable.

\begin{figure}[!tb]
    \centering
    \includegraphics[width=0.97\linewidth]{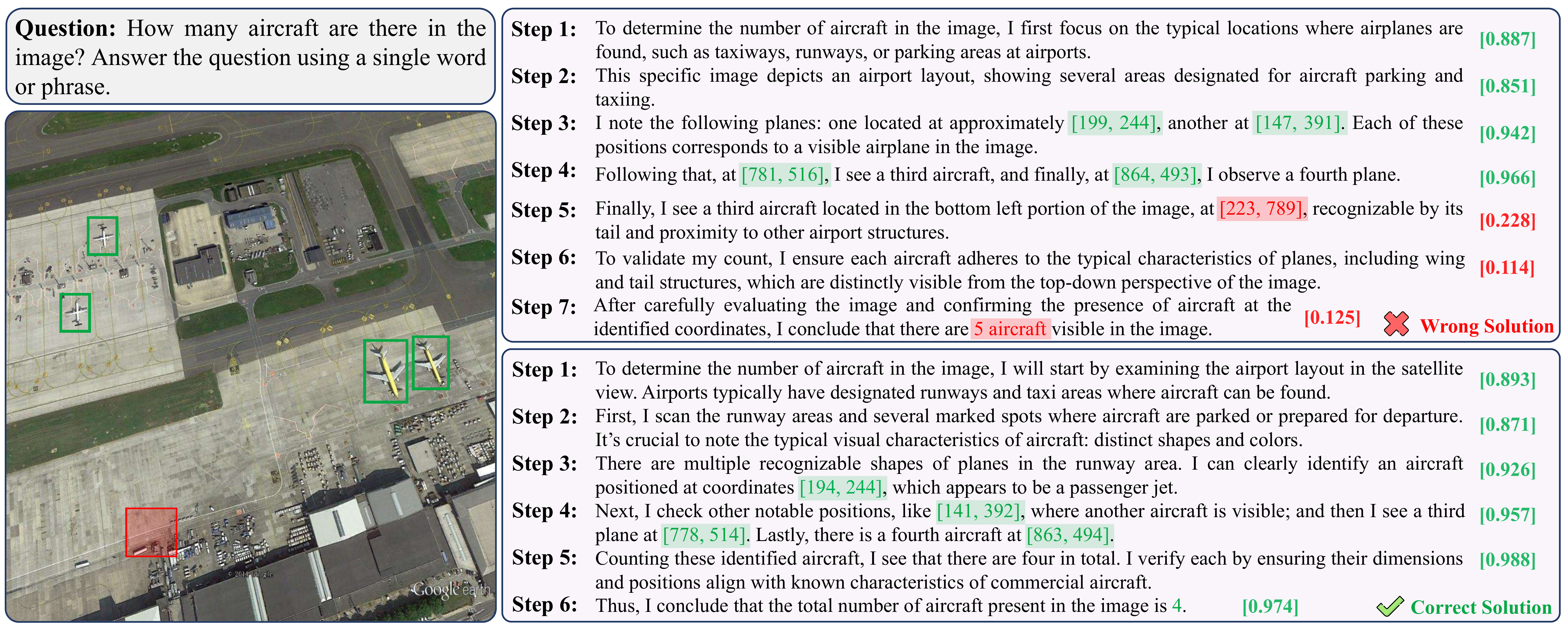}
    \vspace{-0.3cm}
    \caption{Qualitative example of GeoPRM verification. We visualize two reasoning trajectories for an aircraft counting task. In the top erroneous attempt, GeoSolver hallucinates a non-existent aircraft at Step 5, which GeoPRM instantly penalizes with a sharp score drop (from 0.966 to 0.228). The bottom trajectory demonstrates how GeoSolver avoids this error to reach a fully grounded conclusion.}
    \label{fig:vis_quali}
    \vspace{-0.5cm}
\end{figure}

\subsection{Qualitative Analysis}
\label{sec:qualitative}
To intuitively illustrate the system-level impact of our approach, Figure~\ref{fig:vis_quali} demonstrates how process supervision actively intervenes during TTS. While autoregressive generation might blindly commit to a hallucinated object (as seen in the ungrounded Step 5), GeoPRM translates this subtle visual-textual misalignment into a measurable drop-moment. By integrating these dense, token-level signals, GeoSolver effectively prunes deceptive branches early in the search space, ensuring that the final output is strictly anchored to the visual evidence rather than logical extrapolation.

\section{Conclusion}
\label{sec:conclusion}

In this paper, we introduce GeoSolver, a comprehensive framework that scales geospatial reasoning through rigorous process supervision. To combat spatial hallucinations, we construct the Geo-PRM-2M dataset by leveraging Entropy-Guided MCTS and Synthetic Hallucination Injection, enabling the training of GeoPRM as a highly calibrated token-level verifier. For policy alignment, we propose Process-Aware Tree-GRPO, an algorithm that synergizes entropy-guided tree exploration with a process-level penalty to effectively mitigate reward hacking and inherent length bias. Consequently, GeoSolver establishes new state-of-the-art benchmarks across diverse remote sensing tasks under standard inference. Furthermore, we demonstrate that GeoPRM unlocks profound Test-Time Scaling capabilities; it not only consistently enhances GeoSolver's performance but also generalizes as a plug-and-play verifier for generalist VLMs. Ultimately, this work establishes a new paradigm for scaling test-time compute in remote sensing, demonstrating that process-guided exploration is essential for achieving verifiable and faithful geospatial intelligence.


%
%
\bibliographystyle{splncs04}
\bibliography{main}

@String(CVPR  = {IEEE Conf. Comput. Vis. Pattern Recog.})

@String(AAAI  = {AAAI})

@String(CVPR  = {CVPR})

@article{Qwen2.5-VL,
  title={Qwen2.5-VL Technical Report},
  author={Bai, Shuai and Chen, Keqin and Liu, Xuejing and Wang, Jialin and Ge, Wenbin and Song, Sibo and Dang, Kai and Wang, Peng and Wang, Shijie and Tang, Jun and Zhong, Humen and Zhu, Yuanzhi and Yang, Mingkun and Li, Zhaohai and Wan, Jianqiang and Wang, Pengfei and Ding, Wei and Fu, Zheren and Xu, Yiheng and Ye, Jiabo and Zhang, Xi and Xie, Tianbao and Cheng, Zesen and Zhang, Hang and Yang, Zhibo and Xu, Haiyang and Lin, Junyang},
  journal={arXiv preprint arXiv:2502.13923},
  year={2025}
}

@misc{deepseekai2025deepseekr1incentivizingreasoningcapability,
      title={DeepSeek-R1: Incentivizing Reasoning Capability in LLMs via Reinforcement Learning}, 
      author={DeepSeek-AI},
      year={2025},
      eprint={2501.12948},
      archivePrefix={arXiv},
      primaryClass={cs.CL},
      url={https://arxiv.org/abs/2501.12948}, 
}

@article{zhu2023minigpt,
  title   = {Minigpt-4: Enhancing vision-language understanding with advanced large language models},
  author  = {Zhu, Deyao and Chen, Jun and Shen, Xiaoqian and Li, Xiang and Elhoseiny, Mohamed},
  journal = {arXiv preprint arXiv:2304.10592},
  year    = {2023}
}

@article{shao2024deepseekmath,
  title={Deepseekmath: Pushing the limits of mathematical reasoning in open language models},
  author={Shao, Zhihong and Wang, Peiyi and Zhu, Qihao and Xu, Runxin and Song, Junxiao and Bi, Xiao and Zhang, Haowei and Zhang, Mingchuan and Li, YK and Wu, Y and others},
  journal={arXiv preprint arXiv:2402.03300},
  year={2024}
}

@article{peng2024multimath,
  title={Multimath: Bridging visual and mathematical reasoning for large language models},
  author={Peng, Shuai and Fu, Di and Gao, Liangcai and Zhong, Xiuqin and Fu, Hongguang and Tang, Zhi},
  journal={arXiv preprint arXiv:2409.00147},
  year={2024}
}

@misc{luo2024skysensegpt,
      title={SkySenseGPT: A Fine-Grained Instruction Tuning Dataset and Model for Remote Sensing Vision-Language Understanding}, 
      author={Junwei Luo and Zhen Pang and Yongjun Zhang and Tingzhu Wang and Linlin Wang and Bo Dang and Jiangwei Lao and Jian Wang and Jingdong Chen and Yihua Tan and Yansheng Li},
      year={2024},
      eprint={2406.10100},
      archivePrefix={arXiv},
      primaryClass={cs.CV},
      url={https://arxiv.org/abs/2406.10100}, 
}

@inproceedings{kuckreja2024geochat,
  title={Geochat: Grounded large vision-language model for remote sensing},
  author={Kuckreja, Kartik and Danish, Muhammad Sohail and Naseer, Muzammal and Das, Abhijit and Khan, Salman and Khan, Fahad Shahbaz},
  booktitle={Proceedings of the IEEE/CVF Conference on Computer Vision and Pattern Recognition},
  pages={27831--27840},
  year={2024}
}

@inproceedings{pang2025vhm,
  title={Vhm: Versatile and honest vision language model for remote sensing image analysis},
  author={Pang, Chao and Weng, Xingxing and Wu, Jiang and Li, Jiayu and Liu, Yi and Sun, Jiaxing and Li, Weijia and Wang, Shuai and Feng, Litong and Xia, Gui-Song and others},
  booktitle={Proceedings of the AAAI Conference on Artificial Intelligence},
  volume={39},
  number={6},
  pages={6381--6388},
  year={2025}
}

@article{zhang2024earthmarker,
  title={Earthmarker: A visual prompting multi-modal large language model for remote sensing},
  author={Zhang, Wei and Cai, Miaoxin and Zhang, Tong and Zhuang, Yin and Li, Jun and Mao, Xuerui},
  journal={IEEE Transactions on Geoscience and Remote Sensing},
  year={2024},
  publisher={IEEE}
}

@misc{bai2025qwen2,
      title={Qwen2.5-VL Technical Report}, 
      author={Shuai Bai and Keqin Chen and Xuejing Liu and Jialin Wang and Wenbin Ge and Sibo Song and Kai Dang and Peng Wang and Shijie Wang and Jun Tang and Humen Zhong and Yuanzhi Zhu and Mingkun Yang and Zhaohai Li and Jianqiang Wan and Pengfei Wang and Wei Ding and Zheren Fu and Yiheng Xu and Jiabo Ye and Xi Zhang and Tianbao Xie and Zesen Cheng and Hang Zhang and Zhibo Yang and Haiyang Xu and Junyang Lin},
      year={2025},
      eprint={2502.13923},
      archivePrefix={arXiv},
      primaryClass={cs.CV},
      url={https://arxiv.org/abs/2502.13923}, 
}

@misc{chenminigpt,
      title={MiniGPT-v2: large language model as a unified interface for vision-language multi-task learning}, 
      author={Jun Chen and Deyao Zhu and Xiaoqian Shen and Xiang Li and Zechun Liu and Pengchuan Zhang and Raghuraman Krishnamoorthi and Vikas Chandra and Yunyang Xiong and Mohamed Elhoseiny},
      year={2023},
      eprint={2310.09478},
      archivePrefix={arXiv},
        primaryClass={cs.CV},
      url={https://arxiv.org/abs/2310.09478}, 
}

@article{zhang2024earthgpt,
  title={Earthgpt: A universal multi-modal large language model for multi-sensor image comprehension in remote sensing domain},
  author={Zhang, Wei and Cai, Miaoxin and Zhang, Tong and Zhuang, Yin and Mao, Xuerui},
  journal={IEEE Transactions on Geoscience and Remote Sensing},
  year={2024},
  publisher={IEEE}
}

@inproceedings{xia2018dota,
  title={DOTA: A large-scale dataset for object detection in aerial images},
  author={Xia, Gui-Song and Bai, Xiang and Ding, Jian and Zhu, Zhen and Belongie, Serge and Luo, Jiebo and Datcu, Mihai and Pelillo, Marcello and Zhang, Liangpei},
  booktitle={Proceedings of the IEEE conference on computer vision and pattern recognition},
  pages={3974--3983},
  year={2018}
}

@ARTICLE{9560031,
  author={Ding, Jian and Xue, Nan and Xia, Gui-Song and Bai, Xiang and Yang, Wen and Yang, Michael and Belongie, Serge and Luo, Jiebo and Datcu, Mihai and Pelillo, Marcello and Zhang, Liangpei},
  journal={IEEE Transactions on Pattern Analysis and Machine Intelligence},
  title={Object Detection in Aerial Images: A Large-Scale Benchmark and Challenges},
  year={2021},
  volume={},
  number={},
  pages={1-1},
  doi={10.1109/TPAMI.2021.3117983}}

@InProceedings{Ding_2019_CVPR,
author = {Jian, Ding and Nan, Xue and Yang, Long and Gui-Song, Xia and Qikai Lu},
title = {Learning RoI Transformer for Detecting Oriented Objects in Aerial Images},
booktitle = {The IEEE Conference on Computer Vision and Pattern Recognition (CVPR)},
month = {June},
year = {2019}
}

@article{sun2022fair1m,
  title={FAIR1M: A benchmark dataset for fine-grained object recognition in high-resolution remote sensing imagery},
  author={Sun, Xian and Wang, Peijin and Yan, Zhiyuan and Xu, Feng and Wang, Ruiping and Diao, Wenhui and Chen, Jin and Li, Jihao and Feng, Yingchao and Xu, Tao and others},
  journal={ISPRS Journal of Photogrammetry and Remote Sensing},
  volume={184},
  pages={116--130},
  year={2022},
  publisher={Elsevier}
}

@article{lobry2020rsvqa,
  title={RSVQA: Visual question answering for remote sensing data},
  author={Lobry, Sylvain and Marcos, Diego and Murray, Jesse and Tuia, Devis},
  journal={IEEE Transactions on Geoscience and Remote Sensing},
  volume={58},
  number={12},
  pages={8555--8566},
  year={2020},
  publisher={IEEE}
}

@article{lu2017exploring,
  title={Exploring models and data for remote sensing image caption generation},
  author={Lu, Xiaoqiang and Wang, Binqiang and Zheng, Xiangtao and Li, Xuelong},
  journal={IEEE Transactions on Geoscience and Remote Sensing},
  volume={56},
  number={4},
  pages={2183--2195},
  year={2017},
  publisher={IEEE}
}

@article{yuan2021exploring,
  title={Exploring a Fine-Grained Multiscale Method for Cross-Modal Remote Sensing Image Retrieval},
  author={Yuan, Zhiqiang and Zhang, Wenkai and Fu, Kun and Li, Xuan and Deng, Chubo and Wang, Hongqi and Sun, Xian},
  journal={IEEE Transactions on Geoscience and Remote Sensing},
  volume={60},
  pages={1--19},
  year={2021},
  publisher={IEEE}
}

@article{cheng2022nwpu,
  title={NWPU-captions dataset and MLCA-net for remote sensing image captioning},
  author={Cheng, Qimin and Huang, Haiyan and Xu, Yuan and Zhou, Yuzhuo and Li, Huanying and Wang, Zhongyuan},
  journal={IEEE Transactions on Geoscience and Remote Sensing},
  volume={60},
  pages={1--19},
  year={2022},
  publisher={IEEE}
}

@article{zhan2023rsvg,
  title={Rsvg: Exploring data and models for visual grounding on remote sensing data},
  author={Zhan, Yang and Xiong, Zhitong and Yuan, Yuan},
  journal={IEEE Transactions on Geoscience and Remote Sensing},
  volume={61},
  pages={1--13},
  year={2023},
  publisher={IEEE}
}

@inproceedings{liu2024rotated,
  title={Rotated multi-scale interaction network for referring remote sensing image segmentation},
  author={Liu, Sihan and Ma, Yiwei and Zhang, Xiaoqing and Wang, Haowei and Ji, Jiayi and Sun, Xiaoshuai and Ji, Rongrong},
  booktitle={Proceedings of the IEEE/CVF Conference on Computer Vision and Pattern Recognition},
  pages={26658--26668},
  year={2024}
}

@inproceedings{sun2022visual,
  title={Visual grounding in remote sensing images},
  author={Sun, Yuxi and Feng, Shanshan and Li, Xutao and Ye, Yunming and Kang, Jian and Huang, Xu},
  booktitle={Proceedings of the 30th ACM International conference on Multimedia},
  pages={404--412},
  year={2022}
}

@inproceedings{NEURIPS2024_05b7f821,
 author = {Li, Xiang and Ding, Jian and Elhoseiny, Mohamed},
 booktitle = {Advances in Neural Information Processing Systems},
 pages = {3229--3242},
 publisher = {Curran Associates, Inc.},
 title = {VRSBench: A Versatile Vision-Language Benchmark Dataset for Remote Sensing Image Understanding},
 volume = {37},
 year = {2024}
}

@article{long2017accurate,
  title={Accurate object localization in remote sensing images based on convolutional neural networks},
  author={Long, Yang and Gong, Yiping and Xiao, Zhifeng and Liu, Qing},
  journal={IEEE Transactions on Geoscience and Remote Sensing},
  volume={55},
  number={5},
  pages={2486--2498},
  year={2017},
  publisher={IEEE}
}

@article{zhang2019hierarchical,
  title={Hierarchical and robust convolutional neural network for very high-resolution remote sensing object detection},
  author={Zhang, Yuanlin and Yuan, Yuan and Feng, Yachuang and Lu, Xiaoqiang},
  journal={IEEE Transactions on Geoscience and Remote Sensing},
  volume={57},
  number={8},
  pages={5535--5548},
  year={2019},
  publisher={IEEE}
}

@article{cheng2014multi,
  title={Multi-class geospatial object detection and geographic image classification based on collection of part detectors},
  author={Cheng, Gong and Han, Junwei and Zhou, Peicheng and Guo, Lei},
  journal={ISPRS Journal of Photogrammetry and Remote Sensing},
  volume={98},
  pages={119--132},
  year={2014},
  publisher={Elsevier}
}

@article{cheng2017remote,
  title={Remote sensing image scene classification: Benchmark and state of the art},
  author={Cheng, Gong and Han, Junwei and Lu, Xiaoqiang},
  journal={Proceedings of the IEEE},
  volume={105},
  number={10},
  pages={1865--1883},
  year={2017},
  publisher={IEEE}
}

@article{xia2017aid,
  title={AID: A benchmark data set for performance evaluation of aerial scene classification},
  author={Xia, Gui-Song and Hu, Jingwen and Hu, Fan and Shi, Baoguang and Bai, Xiang and Zhong, Yanfei and Zhang, Liangpei and Lu, Xiaoqiang},
  journal={IEEE Transactions on Geoscience and Remote Sensing},
  volume={55},
  number={7},
  pages={3965--3981},
  year={2017},
  publisher={IEEE}
}

@inproceedings{xia2010structural,
  title={Structural high-resolution satellite image indexing},
  author={Xia, Gui-Song and Yang, Wen and Delon, Julie and Gousseau, Yann and Sun, Hong and Ma{\^\i}tre, Henri},
  booktitle={ISPRS TC VII Symposium-100 Years ISPRS},
  volume={38},
  pages={298--303},
  year={2010}
}

@inproceedings{soni2025earthdial,
  title={Earthdial: Turning multi-sensory earth observations to interactive dialogues},
  author={Soni, Sagar and Dudhane, Akshay and Debary, Hiyam and Fiaz, Mustansar and Munir, Muhammad Akhtar and Danish, Muhammad Sohail and Fraccaro, Paolo and Watson, Campbell D and Klein, Levente J and Khan, Fahad Shahbaz and others},
  booktitle={Proceedings of the Computer Vision and Pattern Recognition Conference},
  pages={14303--14313},
  year={2025}
}

@article{li2025segearth,
  title={Segearth-r1: Geospatial pixel reasoning via large language model},
  author={Li, Kaiyu and Xin, Zepeng and Pang, Li and Pang, Chao and Deng, Yupeng and Yao, Jing and Xia, Guisong and Meng, Deyu and Wang, Zhi and Cao, Xiangyong},
  journal={arXiv preprint arXiv:2504.09644},
  year={2025}
}

@article{yao2025remotereasoner,
  title={RemoteReasoner: Towards Unifying Geospatial Reasoning Workflow},
  author={Yao, Liang and Liu, Fan and Lu, Hongbo and Zhang, Chuanyi and Min, Rui and Xu, Shengxiang and Di, Shimin and Peng, Pai},
  journal={arXiv preprint arXiv:2507.19280},
  year={2025}
}

@InProceedings{Zhu_2025_CVPR_skysense_o,
    author    = {Zhu, Qi and Lao, Jiangwei and Ji, Deyi and Luo, Junwei and Wu, Kang and Zhang, Yingying and Ru, Lixiang and Wang, Jian and Chen, Jingdong and Yang, Ming and Liu, Dong and Zhao, Feng},
    title     = {SkySense-O: Towards Open-World Remote Sensing Interpretation with Vision-Centric Visual-Language Modeling},
    booktitle = {Proceedings of the Computer Vision and Pattern Recognition Conference (CVPR)},
    month     = {June},
    year      = {2025},
    pages     = {14733-14744}
}

@inproceedings{guo2024skysense,
    author    = {Guo, Xin and Lao, Jiangwei and Dang, Bo and Zhang, Yingying and Yu, Lei and Ru, Lixiang and Zhong, Liheng and Huang, Ziyuan and Wu, Kang and Hu, Dingxiang and He, Huimei and Wang, Jian and Chen, Jingdong and Yang, Ming and Zhang, Yongjun and Li, Yansheng},
    title     = {SkySense: A Multi-Modal Remote Sensing Foundation Model Towards Universal Interpretation for Earth Observation Imagery},
    booktitle = {Proceedings of the IEEE/CVF Conference on Computer Vision and Pattern Recognition (CVPR)},
    month     = {June},
    year      = {2024},
    pages     = {27672-27683}
}

@article{hu2025ringmo,
  title={RingMo-Agent: A Unified Remote Sensing Foundation Model for Multi-Platform and Multi-Modal Reasoning},
  author={Hu, Huiyang and Wang, Peijin and Feng, Yingchao and Wei, Kaiwen and Yin, Wenxin and Diao, Wenhui and Wang, Mengyu and Bi, Hanbo and Kang, Kaiyue and Ling, Tong and others},
  journal={arXiv preprint arXiv:2507.20776},
  year={2025}
}

@article{zhao2016dirichlet,
  title={Dirichlet-derived multiple topic scene classification model fusing heterogeneous features for high spatial resolution remote sensing imagery},
  author={Zhao, Bei and Zhong, Yanfei and Xia, GS and Zhang, Liangpei},
  journal={IEEE Trans. Geosci. Remote Sens},
  volume={54},
  number={4},
  pages={2108--2123},
  year={2016}
}

@article{zhao2016fisher,
  title={The Fisher kernel coding framework for high spatial resolution scene classification},
  author={Zhao, Bei and Zhong, Yanfei and Zhang, Liangpei and Huang, Bo},
  journal={Remote Sensing},
  volume={8},
  number={2},
  pages={157},
  year={2016},
  publisher={MDPI}
}

@article{zhu2016bag,
  title={Bag-of-visual-words scene classifier with local and global features for high spatial resolution remote sensing imagery},
  author={Zhu, Qiqi and Zhong, Yanfei and Zhao, Bei and Xia, Gui-Song and Zhang, Liangpei},
  journal={IEEE Geoscience and Remote Sensing Letters},
  volume={13},
  number={6},
  pages={747--751},
  year={2016},
  publisher={IEEE}
}

@InProceedings{Nilsback08,
   author = "Yang, Yi and Newsam, Shawn",
   title = "Bag-Of-Visual-Words and Spatial Extensions for Land-Use Classification",
   booktitle = "ACM SIGSPATIAL International Conference on Advances in Geographic Information Systems (ACM GIS)",
   year = "2010",
}

@misc{openai2025gpt5_blog,
  title        = {Introducing GPT-5},
  author       = {OpenAI},
  year         = {2025},
  howpublished = {\url{https://openai.com/introducing-gpt-5/}},
}

@article{comanici2025gemini,
  title={Gemini 2.5: Pushing the frontier with advanced reasoning, multimodality, long context, and next generation agentic capabilities},
  author={Comanici, Gheorghe and Bieber, Eric and Schaekermann, Mike and Pasupat, Ice and Sachdeva, Noveen and Dhillon, Inderjit and Blistein, Marcel and Ram, Ori and Zhang, Dan and Rosen, Evan and others},
  journal={arXiv preprint arXiv:2507.06261},
  year={2025}
}

@misc{anthropic2025claude4_opus_sonnet,
  title        = {Claude Opus 4 \& Claude Sonnet 4 System Card},
  author       = {Anthropic},
  year         = {2025},
  howpublished = {\url{https://www.anthropic.com/claude-4-system-card}},
}

@misc{vteam2025glm45vglm41vthinkingversatilemultimodal,
      title={GLM-4.5V and GLM-4.1V-Thinking: Towards Versatile Multimodal Reasoning with Scalable Reinforcement Learning}, 
      author={V Team and Wenyi Hong and Wenmeng Yu and Xiaotao Gu and Guo Wang and Guobing Gan and Haomiao Tang and Jiale Cheng and Ji Qi and Junhui Ji and Lihang Pan and Shuaiqi Duan and Weihan Wang and Yan Wang and Yean Cheng and Zehai He and Zhe Su and Zhen Yang and Ziyang Pan and Aohan Zeng and Baoxu Wang and Bin Chen and Boyan Shi and Changyu Pang and Chenhui Zhang and Da Yin and Fan Yang and Guoqing Chen and Jiazheng Xu and Jiale Zhu and Jiali Chen and Jing Chen and Jinhao Chen and Jinghao Lin and Jinjiang Wang and Junjie Chen and Leqi Lei and Letian Gong and Leyi Pan and Mingdao Liu and Mingde Xu and Mingzhi Zhang and Qinkai Zheng and Sheng Yang and Shi Zhong and Shiyu Huang and Shuyuan Zhao and Siyan Xue and Shangqin Tu and Shengbiao Meng and Tianshu Zhang and Tianwei Luo and Tianxiang Hao and Tianyu Tong and Wenkai Li and Wei Jia and Xiao Liu and Xiaohan Zhang and Xin Lyu and Xinyue Fan and Xuancheng Huang and Yanling Wang and Yadong Xue and Yanfeng Wang and Yanzi Wang and Yifan An and Yifan Du and Yiming Shi and Yiheng Huang and Yilin Niu and Yuan Wang and Yuanchang Yue and Yuchen Li and Yutao Zhang and Yuting Wang and Yu Wang and Yuxuan Zhang and Zhao Xue and Zhenyu Hou and Zhengxiao Du and Zihan Wang and Peng Zhang and Debing Liu and Bin Xu and Juanzi Li and Minlie Huang and Yuxiao Dong and Jie Tang},
      year={2025},
      eprint={2507.01006},
      archivePrefix={arXiv},
      primaryClass={cs.CV},
      url={https://arxiv.org/abs/2507.01006}, 
}

@misc{kimiteam2025kimivltechnicalreport,
      title={{Kimi-VL} Technical Report}, 
      author={Kimi Team and Angang Du and Bohong Yin and Bowei Xing and Bowen Qu and Bowen Wang and Cheng Chen and Chenlin Zhang and Chenzhuang Du and Chu Wei and Congcong Wang and Dehao Zhang and Dikang Du and Dongliang Wang and Enming Yuan and Enzhe Lu and Fang Li and Flood Sung and Guangda Wei and Guokun Lai and Han Zhu and Hao Ding and Hao Hu and Hao Yang and Hao Zhang and Haoning Wu and Haotian Yao and Haoyu Lu and Heng Wang and Hongcheng Gao and Huabin Zheng and Jiaming Li and Jianlin Su and Jianzhou Wang and Jiaqi Deng and Jiezhong Qiu and Jin Xie and Jinhong Wang and Jingyuan Liu and Junjie Yan and Kun Ouyang and Liang Chen and Lin Sui and Longhui Yu and Mengfan Dong and Mengnan Dong and Nuo Xu and Pengyu Cheng and Qizheng Gu and Runjie Zhou and Shaowei Liu and Sihan Cao and Tao Yu and Tianhui Song and Tongtong Bai and Wei Song and Weiran He and Weixiao Huang and Weixin Xu and Xiaokun Yuan and Xingcheng Yao and Xingzhe Wu and Xinxing Zu and Xinyu Zhou and Xinyuan Wang and Y. Charles and Yan Zhong and Yang Li and Yangyang Hu and Yanru Chen and Yejie Wang and Yibo Liu and Yibo Miao and Yidao Qin and Yimin Chen and Yiping Bao and Yiqin Wang and Yongsheng Kang and Yuanxin Liu and Yulun Du and Yuxin Wu and Yuzhi Wang and Yuzi Yan and Zaida Zhou and Zhaowei Li and Zhejun Jiang and Zheng Zhang and Zhilin Yang and Zhiqi Huang and Zihao Huang and Zijia Zhao and Ziwei Chen},
      year={2025},
      eprint={2504.07491},
      archivePrefix={arXiv},
      primaryClass={cs.CV},
      url={https://arxiv.org/abs/2504.07491}, 
}

@inproceedings{fini2025multimodal,
  title={Multimodal autoregressive pre-training of large vision encoders},
  author={Fini, Enrico and Shukor, Mustafa and Li, Xiujun and Dufter, Philipp and Klein, Michal and Haldimann, David and Aitharaju, Sai and da Costa, Victor G Turrisi and B{\'e}thune, Louis and Gan, Zhe and others},
  booktitle={Proceedings of the Computer Vision and Pattern Recognition Conference},
  pages={9641--9654},
  year={2025}
}

@article{Qwen3-VL,
      title={Qwen3-VL Technical Report}, 
      author={Shuai Bai and Yuxuan Cai and Ruizhe Chen and Keqin Chen and Xionghui Chen and Zesen Cheng and Lianghao Deng and Wei Ding and Chang Gao and Chunjiang Ge and Wenbin Ge and Zhifang Guo and Qidong Huang and Jie Huang and Fei Huang and Binyuan Hui and Shutong Jiang and Zhaohai Li and Mingsheng Li and Mei Li and Kaixin Li and Zicheng Lin and Junyang Lin and Xuejing Liu and Jiawei Liu and Chenglong Liu and Yang Liu and Dayiheng Liu and Shixuan Liu and Dunjie Lu and Ruilin Luo and Chenxu Lv and Rui Men and Lingchen Meng and Xuancheng Ren and Xingzhang Ren and Sibo Song and Yuchong Sun and Jun Tang and Jianhong Tu and Jianqiang Wan and Peng Wang and Pengfei Wang and Qiuyue Wang and Yuxuan Wang and Tianbao Xie and Yiheng Xu and Haiyang Xu and Jin Xu and Zhibo Yang and Mingkun Yang and Jianxin Yang and An Yang and Bowen Yu and Fei Zhang and Hang Zhang and Xi Zhang and Bo Zheng and Humen Zhong and Jingren Zhou and Fan Zhou and Jing Zhou and Yuanzhi Zhu and Ke Zhu},
	  journal={arXiv preprint arXiv:2511.21631},
      year={2025}
}

@article{liu2025towardsrsthinker,
  title={Towards Faithful Reasoning in Remote Sensing: A Perceptually-Grounded GeoSpatial Chain-of-Thought for Vision-Language Models},
  author={Liu, Jiaqi and Sun, Lang and Fu, Ronghao and Yang, Bo},
  journal={arXiv preprint arXiv:2509.22221},
  year={2025}
}

@article{liu2025skymoe,
  title={SkyMoE: A Vision-Language Foundation Model for Enhancing Geospatial Interpretation with Mixture of Experts},
  author={Liu, Jiaqi and Fu, Ronghao and Sun, Lang and Liu, Haoran and Yang, Xiao and Zhang, Weipeng and Na, Xu and Duan, Zhuoran and Yang, Bo},
  journal={arXiv preprint arXiv:2512.02517},
  year={2025}
}

@article{liu2025geodit,
  title={GeoDiT: A Diffusion-based Vision-Language Model for Geospatial Understanding},
  author={Liu, Jiaqi and Fu, Ronghao and Liu, Haoran and Sun, Lang and Yang, Bo},
  journal={arXiv preprint arXiv:2512.02505},
  year={2025}
}

@article{fiaz2025geovlmr1,
  title={Geovlm-r1: Reinforcement fine-tuning for improved remote sensing reasoning},
  author={Fiaz, Mustansar and Debary, Hiyam and Fraccaro, Paolo and Paudel, Danda and Van Gool, Luc and Khan, Fahad and Khan, Salman},
  journal={arXiv preprint arXiv:2509.25026},
  year={2025}
}

@article{zhang2025geor1_fewshot,
  title={Geo-R1: Improving Few-Shot Geospatial Referring Expression Understanding with Reinforcement Fine-Tuning},
  author={Zhang, Zilun and Guan, Zian and Zhao, Tiancheng and Shen, Haozhan and Li, Tianyu and Cai, Yuxiang and Su, Zhonggen and Liu, Zhaojun and Yin, Jianwei and Li, Xiang},
  journal={arXiv preprint arXiv:2509.21976},
  year={2025}
}

@article{wang2025geozero,
  title={GeoZero: Incentivizing Reasoning from Scratch on Geospatial Scenes},
  author={Wang, Di and Liu, Shunyu and Jiang, Wentao and Wang, Fengxiang and Liu, Yi and Qin, Xiaolei and Luo, Zhiming and Zhou, Chaoyang and Guo, Haonan and Zhang, Jing and others},
  journal={arXiv preprint arXiv:2511.22645},
  year={2025}
}

@article{shao2025asking_geoeot,
  title={Asking like Socrates: Socrates helps VLMs understand remote sensing images},
  author={Shao, Run and Li, Ziyu and Zhang, Zhaoyang and Xu, Linrui and He, Xinran and Yuan, Hongyuan and He, Bolei and Dai, Yongxing and Yan, Yiming and Chen, Yijun and others},
  journal={arXiv preprint arXiv:2511.22396},
  year={2025}
}

@inproceedings{shabbir2025geopixel,
  title={GeoPixel: Pixel Grounding Large Multimodal Model in Remote Sensing},
  author={Shabbir, Akashah and Zumri, Mohammed and Bennamoun, Mohammed and Khan, Fahad Shahbaz and Khan, Salman},
  booktitle={International Conference on Machine Learning},
  pages={54095--54111},
  year={2025},
  organization={PMLR}
}

@inproceedings{xie2025outcomes_prms,
  title={From outcomes to processes: Guiding PRM learning from ORM for inference-time alignment},
  author={Xie, Bin and Xu, Bingbing and Yuan, Yige and Zhu, Shengmao and Shen, Huawei},
  booktitle={Proceedings of the 63rd Annual Meeting of the Association for Computational Linguistics (Volume 1: Long Papers)},
  pages={19291--19307},
  year={2025}
}

@article{zheng2025survey_prm,
  title={A survey of process reward models: From outcome signals to process supervisions for large language models},
  author={Zheng, Congming and Zhu, Jiachen and Ou, Zhuoying and Chen, Yuxiang and Zhang, Kangning and Shan, Rong and Zheng, Zeyu and Yang, Mengyue and Lin, Jianghao and Yu, Yong and others},
  journal={arXiv preprint arXiv:2510.08049},
  year={2025}
}

@article{li2025system_prm,
  title={From system 1 to system 2: A survey of reasoning large language models},
  author={Li, Zhong-Zhi and Zhang, Duzhen and Zhang, Ming-Liang and Zhang, Jiaxin and Liu, Zengyan and Yao, Yuxuan and Xu, Haotian and Zheng, Junhao and Wang, Pei-Jie and Chen, Xiuyi and others},
  journal={arXiv preprint arXiv:2502.17419},
  year={2025}
}

@article{luo2025ursa,
  title={Ursa: Understanding and verifying chain-of-thought reasoning in multimodal mathematics},
  author={Luo, Ruilin and Zheng, Zhuofan and Wang, Yifan and Yu, Yiyao and Ni, Xinzhe and Lin, Zicheng and Zeng, Jin and Yang, Yujiu},
  journal={arXiv e-prints},
  pages={arXiv--2501},
  year={2025}
}

@article{yang2024qwen2_5math,
  title={Qwen2. 5-math technical report: Toward mathematical expert model via self-improvement},
  author={Yang, An and Zhang, Beichen and Hui, Binyuan and Gao, Bofei and Yu, Bowen and Li, Chengpeng and Liu, Dayiheng and Tu, Jianhong and Zhou, Jingren and Lin, Junyang and others},
  journal={arXiv preprint arXiv:2409.12122},
  year={2024}
}

@inproceedings{zhuang2025mathpuma,
  title={Math-puma: Progressive upward multimodal alignment to enhance mathematical reasoning},
  author={Zhuang, Wenwen and Huang, Xin and Zhang, Xiantao and Zeng, Jin},
  booktitle={Proceedings of the AAAI Conference on Artificial Intelligence},
  volume={39},
  number={24},
  pages={26183--26191},
  year={2025}
}

@inproceedings{yu2025chainmath,
  title={Chain-of-reasoning: Towards unified mathematical reasoning in large language models via a multi-paradigm perspective},
  author={Yu, Yiyao and Zhang, Yuxiang and Zhang, Dongdong and Liang, Xiao and Zhang, Hengyuan and Zhang, Xingxing and Khademi, Mahmoud and Awadalla, Hany Hassan and Wang, Junjie and Yang, Yujiu and others},
  booktitle={Proceedings of the 63rd Annual Meeting of the Association for Computational Linguistics (Volume 1: Long Papers)},
  pages={24914--24937},
  year={2025}
}

@article{huang2025visionr1_math,
  title={Vision-r1: Incentivizing reasoning capability in multimodal large language models},
  author={Huang, Wenxuan and Jia, Bohan and Zhai, Zijie and Cao, Shaosheng and Ye, Zheyu and Zhao, Fei and Xu, Zhe and Hu, Yao and Lin, Shaohui},
  journal={arXiv preprint arXiv:2503.06749},
  year={2025}
}

@article{zhang2024generative_tts,
  title={Generative verifiers: Reward modeling as next-token prediction},
  author={Zhang, Lunjun and Hosseini, Arian and Bansal, Hritik and Kazemi, Mehran and Kumar, Aviral and Agarwal, Rishabh},
  journal={arXiv preprint arXiv:2408.15240},
  year={2024}
}

@inproceedings{muennighoff2025s1_tts,
  title={s1: Simple test-time scaling},
  author={Muennighoff, Niklas and Yang, Zitong and Shi, Weijia and Li, Xiang Lisa and Fei-Fei, Li and Hajishirzi, Hannaneh and Zettlemoyer, Luke and Liang, Percy and Cand{\`e}s, Emmanuel and Hashimoto, Tatsunori B},
  booktitle={Proceedings of the 2025 Conference on Empirical Methods in Natural Language Processing},
  pages={20286--20332},
  year={2025}
}

@inproceedings{peng2025rewarding_graphprm,
  title={Rewarding graph reasoning process makes llms more generalized reasoners},
  author={Peng, Miao and Chen, Nuo and Suo, Zongrui and Li, Jia},
  booktitle={Proceedings of the 31st ACM SIGKDD Conference on Knowledge Discovery and Data Mining V. 2},
  pages={2257--2268},
  year={2025}
}

@article{luo2024improve_mcts,
  title={Improve mathematical reasoning in language models by automated process supervision},
  author={Luo, Liangchen and Liu, Yinxiao and Liu, Rosanne and Phatale, Samrat and Guo, Meiqi and Lara, Harsh and Li, Yunxuan and Shu, Lei and Zhu, Yun and Meng, Lei and others},
  journal={arXiv preprint arXiv:2406.06592},
  year={2024}
}

@article{wei2022chain_cot,
  title={Chain-of-thought prompting elicits reasoning in large language models},
  author={Wei, Jason and Wang, Xuezhi and Schuurmans, Dale and Bosma, Maarten and Xia, Fei and Chi, Ed and Le, Quoc V and Zhou, Denny and others},
  journal={Advances in neural information processing systems},
  volume={35},
  pages={24824--24837},
  year={2022}
}

@article{cobbe2021training_orm,
  title={Training verifiers to solve math word problems},
  author={Cobbe, Karl and Kosaraju, Vineet and Bavarian, Mohammad and Chen, Mark and Jun, Heewoo and Kaiser, Lukasz and Plappert, Matthias and Tworek, Jerry and Hilton, Jacob and Nakano, Reiichiro and others},
  journal={arXiv preprint arXiv:2110.14168},
  year={2021}
}

@inproceedings{lightman2023let_prm,
  title={Let's verify step by step},
  author={Lightman, Hunter and Kosaraju, Vineet and Burda, Yuri and Edwards, Harrison and Baker, Bowen and Lee, Teddy and Leike, Jan and Schulman, John and Sutskever, Ilya and Cobbe, Karl},
  booktitle={The twelfth international conference on learning representations},
  year={2023}
}

@article{browne2012survey_mcts,
  title={A survey of monte carlo tree search methods},
  author={Browne, Cameron B and Powley, Edward and Whitehouse, Daniel and Lucas, Simon M and Cowling, Peter I and Rohlfshagen, Philipp and Tavener, Stephen and Perez, Diego and Samothrakis, Spyridon and Colton, Simon},
  journal={IEEE Transactions on Computational Intelligence and AI in games},
  volume={4},
  number={1},
  pages={1--43},
  year={2012},
  publisher={IEEE}
}

@article{pak2025correction_remotesense,
  title={Correction of systematic image misalignment in direct georeferencing of UAV multispectral imagery},
  author={Pak, Hui Ying and Lin, Weisi and Law, Adrian Wing-Keung},
  journal={International Journal of Remote Sensing},
  volume={46},
  number={3},
  pages={930--952},
  year={2025},
  publisher={Taylor \& Francis}
}

@article{lenton2024remotely_remotesense,
  title={Remotely sensing potential climate change tipping points across scales},
  author={Lenton, Timothy M and Abrams, Jesse F and Bartsch, Annett and Bathiany, Sebastian and Boulton, Chris A and Buxton, Joshua E and Conversi, Alessandra and Cunliffe, Andrew M and Hebden, Sophie and Lavergne, Thomas and others},
  journal={nature communications},
  volume={15},
  number={1},
  pages={343},
  year={2024},
  publisher={Nature Publishing Group UK London}
}

@article{schulman2017proximal_ppo,
  title={Proximal policy optimization algorithms},
  author={Schulman, John and Wolski, Filip and Dhariwal, Prafulla and Radford, Alec and Klimov, Oleg},
  journal={arXiv preprint arXiv:1707.06347},
  year={2017}
}

@inproceedings{hou2025treerl,
  title={Treerl: Llm reinforcement learning with on-policy tree search},
  author={Hou, Zhenyu and Hu, Ziniu and Li, Yujiang and Lu, Rui and Tang, Jie and Dong, Yuxiao},
  booktitle={Proceedings of the 63rd Annual Meeting of the Association for Computational Linguistics (Volume 1: Long Papers)},
  pages={12355--12369},
  year={2025}
}

@article{guo2025segment_spo,
  title={Segment policy optimization: Effective segment-level credit assignment in rl for large language models},
  author={Guo, Yiran and Xu, Lijie and Liu, Jie and Ye, Dan and Qiu, Shuang},
  journal={arXiv preprint arXiv:2505.23564},
  year={2025}
}

@article{ji2025tree_grpo,
  title={Tree search for llm agent reinforcement learning},
  author={Ji, Yuxiang and Ma, Ziyu and Wang, Yong and Chen, Guanhua and Chu, Xiangxiang and Wu, Liaoni},
  journal={arXiv preprint arXiv:2509.21240},
  year={2025}
}

@inproceedings{gao2023pal,
  title={Pal: Program-aided language models},
  author={Gao, Luyu and Madaan, Aman and Zhou, Shuyan and Alon, Uri and Liu, Pengfei and Yang, Yiming and Callan, Jamie and Neubig, Graham},
  booktitle={International conference on machine learning},
  pages={10764--10799},
  year={2023},
  organization={PMLR}
}

@article{yan2024errorradar,
  title={Errorradar: Benchmarking complex mathematical reasoning of multimodal large language models via error detection},
  author={Yan, Yibo and Wang, Shen and Huo, Jiahao and Li, Hang and Li, Boyan and Su, Jiamin and Gao, Xiong and Zhang, Yi-Fan and Xu, Tianlong and Chu, Zhendong and others},
  journal={arXiv preprint arXiv:2410.04509},
  year={2024}
}

@article{snell2024scaling_beam,
  title={Scaling llm test-time compute optimally can be more effective than scaling model parameters},
  author={Snell, Charlie and Lee, Jaehoon and Xu, Kelvin and Kumar, Aviral},
  journal={arXiv preprint arXiv:2408.03314},
  year={2024}
}

@article{loshchilov2017decoupled_adamw,
  title={Decoupled weight decay regularization},
  author={Loshchilov, Ilya and Hutter, Frank},
  journal={arXiv preprint arXiv:1711.05101},
  year={2017}
}

@article{zhou2018patternnet,
  title={PatternNet: A benchmark dataset for performance evaluation of remote sensing image retrieval},
  author={Zhou, Weixun and Newsam, Shawn and Li, Congmin and Shao, Zhenfeng},
  journal={ISPRS journal of photogrammetry and remote sensing},
  volume={145},
  pages={197--209},
  year={2018},
  publisher={Elsevier}
}

@article{rahnemoonfar2021floodnet,
  title={Floodnet: A high resolution aerial imagery dataset for post flood scene understanding},
  author={Rahnemoonfar, Maryam and Chowdhury, Tashnim and Sarkar, Argho and Varshney, Debvrat and Yari, Masoud and Murphy, Robin Roberson},
  journal={IEEE Access},
  volume={9},
  pages={89644--89654},
  year={2021},
  publisher={IEEE}
}

\clearpage

\appendix



\section{Implementation Details}
\label{supp:implementation}

To facilitate reproducibility, we provide comprehensive implementation details covering the hyperparameters across all alignment stages, the data formatting, and the explicit configurations used for Test-Time Scaling (TTS).

\subsection{Supervised Fine-Tuning (SFT)}
During the SFT stage, we initialize our model from the pretrained GLM-4.1V-9B-Base~\cite{vteam2025glm45vglm41vthinkingversatilemultimodal}. The model is fine-tuned on the Geo-CoT380k~\cite{liu2025towardsrsthinker} dataset for 1 epoch. To enforce a structured cognitive process, all training targets in the dataset are strictly unified into a decoupled reasoning-answering format: \textit{<think> reasoning process</think><answer>final answer</answer>}. This explicitly trains the base model to perform perceptually-grounded Chain-of-Thought (CoT) before predicting the final coordinates or labels.

We employ the AdamW~\cite{loshchilov2017decoupled_adamw} optimizer and a weight decay of 0.05. The learning rate is warmed up to a peak of $2 \times 10^{-5}$ over the first 5\% of training steps, followed by a cosine decay schedule down to $1 \times 10^{-6}$. The global batch size is set to 128, and the maximum sequence length is restricted to 2048 tokens.

\subsection{GeoPRM Construction and Training}
\noindent\textbf{MCTS Data Synthesis.} For the Entropy-Guided MCTS, we dynamically expand the reasoning tree by selecting the top $N = 3$ tokens with the highest entropy at each identified uncertain branching point. From each expanded node, we perform $T = 9$ independent linear rollouts. The tree construction iterates for a maximum of $K = 4$ depth levels. The generation temperature during rollouts is set to 1.2 to encourage exploratory diversity. 

\noindent\textbf{Reward Model Training.} GeoPRM is initialized from the SFT checkpoint, $\pi_{\text{sft}}$, with an appended randomly initialized linear classification head. We train GeoPRM on the combined Geo-PRM-2M dataset for 2 epochs using a learning rate of $1 \times 10^{-5}$ and a global batch size of 256. The token-level binary cross-entropy loss is computed exclusively on the reasoning sequence tokens, masking out the prompt and image tokens. 

\subsection{Process-Aware Tree-GRPO}
In the reinforcement learning stage, the policy model $\pi_\theta$ is optimized using the Process-Aware Tree-GRPO algorithm. The learning rate is set to a strictly lower value of $1 \times 10^{-6}$ with a constant schedule to ensure stable policy updates. For the Vanilla GRPO~\cite{deepseekai2025deepseekr1incentivizingreasoningcapability}, we sample $G = 8$ complete reasoning trajectories per prompt. The clipping ratio for the surrogate objective is set to $\epsilon = 0.2$.

\noindent\textbf{Hyperparameter Ablation for Drop-Moment Penalty.} 

\begin{wraptable}{r}{0.25\textwidth}
\centering
\setlength{\tabcolsep}{8pt}
\resizebox{\linewidth}{!}{
\begin{tabular}{ccc}
\toprule
\textbf{ $\rho$ } & \textbf{ $\gamma$ } & \textbf{Score} \\
\midrule
0.3 & 0.7 & \textbf{70.51} \\
0.3 & 0.3 & 68.21 \\
0.3 & 0.5 & 69.66 \\
0.3 & 0.9 & 70.15 \\
0.2 & 0.7 & 69.06 \\
0.4 & 0.7 & 69.42 \\
\bottomrule
\end{tabular}
}
\vspace{-0.4cm}
\caption{Ablation study of the drop-moment hyperparameters. We report the average score across multiple tasks.}
\label{tab:supp_ablation_penalty}
\vspace{-0.5cm}
\end{wraptable}

\noindent To extract reliable verification signals from GeoPRM, we introduce the drop-moment~\cite{luo2025ursa} penalty mechanism controlled by two critical hyperparameters: the sensitivity threshold ($\rho$) and the penalty factor ($\gamma$). To determine their optimal values, we conduct a grid search ablation study on a held-out validation set, evaluating the average performance across Visual Grounding and Object Counting tasks. 

As shown in Table~\ref{tab:supp_ablation_penalty}, setting $\rho = 0.3$ and $\gamma = 0.7$ yields the best overall performance. A lower threshold ($\rho = 0.2$) makes the penalty too aggressive, mistakenly penalizing minor confidence fluctuations and collapsing the reasoning length. Conversely, a higher threshold ($\rho = 0.4$) fails to capture genuine, subtle visual hallucinations. Similarly, applying a severe absolute penalty ($\gamma = 0.3, 0.5$) disrupts RL training stability by creating extremely sparse rewards, whereas a mild penalty ($\gamma = 0.9$) insufficiently regularizes the reward hacking behavior.

\subsection{Inference and Test-Time Scaling}
For standard inference, we employ greedy decoding (temperature $\tau = 0.0$). To further unleash the reasoning potential during inference, we explore Test-Time Scaling (TTS) via two distinct decoding strategies guided by GeoPRM:
Best-of-N (BoN): We independently sample $N \in \{2, 4, 8, 16, 32, 64, 128\}$ diverse candidate trajectories per query with a temperature of $\tau = 1.0$ and Top-p of 0.9. The trajectory with the highest average token-level process reward from GeoPRM is selected as the final output. Beam Search: We maintain a beam width of $N \in \{2, 4, 8, 16, 32, 64, 128\}$ and keep top $M=N/2$ to expansion. Unlike BoN which evaluates complete sequences post-hoc, Beam Search utilizes GeoPRM as a step-wise verifier to actively score and prune suboptimal branches during the autoregressive generation. This step-level intervention is particularly effective for deep, multi-step logical spatial reasoning where early course-correction is crucial.

\begin{table}[ht]
\centering
\resizebox{0.95\linewidth}{!}{
\begin{tabular}{llllll}
\toprule
 \textbf{Task} & \textbf{Dataset} & \textbf{Stage  } & \textbf{Split  } & \textbf{Samples  } & \textbf{QA Examples} \\
\midrule
\multirow{4}{*}{Visual Grounding} & DIOR-RSVG~\cite{zhan2023rsvg}& SFT/RL\textbf{  } & train/test\textbf{  } & 31k/8k & \multirow{4}{*}{\makecell[l]{\textbf{User}: Where is a slender small train station in the middle? \\ \textbf{Answer}: [280,371,503,420].}} \\
 & VRSBench~\cite{NEURIPS2024_05b7f821} & SFT/RL & train/test & 20k/16k \\
 & RRSIS-D~\cite{liu2024rotated} & $-$ & test & 16k \\
 & RSVG~\cite{sun2022visual} & $-$ & test & 1k \\
\midrule
\multirow{3}{*}{Object Detection} & DOTAv2~\cite{xia2018dota, Ding_2019_CVPR, 9560031} & SFT/RL & train/test & 29k/9k & \multirow{3}{*}{\makecell[l]{\textbf{User}: Detect all plane in the image. \\ \textbf{Answer}: [[63, 70, 28, 138], [45, 120, 50, 185], [226, 148, 307, 203]].}} \\
 & HRRSD~\cite{zhang2019hierarchical} & SFT/RL & train/test & 11k/11k \\
 & FAIR1M~\cite{sun2022fair1m} & RL & train & 6k \\
\midrule
\multirow{5}{*}{Object Counting} & DOTAv2~\cite{xia2018dota, Ding_2019_CVPR, 9560031} & SFT/RL &  train/test & 29k/9k & \multirow{5}{*}{\makecell[l]{\textbf{User}: How many airport are there in the image? Answer the question \\ using a single word or phrase. \\ \textbf{Answer}: 3.}} \\
 & HRRSD~\cite{zhang2019hierarchical} & SFT/RL &  train/test & 11k/11k \\
 & FAIR1M~\cite{sun2022fair1m} & RL & train & 6k \\
 & NWPU-VHR~\cite{cheng2014multi} & $-$ & test & 1k \\
 & RSOD~\cite{long2017accurate} & $-$ & test & 1k \\
\midrule
\multirow{6}{*}{Scene Classification} & RESISC45~\cite{cheng2017remote} & SFT/RL & train & 25k/6k & \multirow{6}{*}{\makecell[l]{\textbf{User}: Which scene does this image belong to? You must choose one of \\Parking, BaseballField, School, Resort, Industrial, ... \\ \textbf{Answer}: Parking.}} \\
 & AID~\cite{xia2017aid} & SFT/RL &  train/test & 8k/2k \\
 & PatternNet~\cite{zhou2018patternnet} & RL &  train & 30k \\
 & WHU-RS19~\cite{xia2010structural} & $-$ & test & 1k  \\
 & SIRI-WHU~\cite{zhao2016dirichlet, zhao2016fisher, zhu2016bag} & $-$ & test & 2k  \\
 & UCMerced~\cite{Nilsback08} & $-$ & test & 2k \\
\midrule
\multirow{3}{*}{Visual Question Answering} & VRSBench~\cite{NEURIPS2024_05b7f821} & SFT/RL & train/test & 20k/16k & \multirow{3}{*}{\makecell[l]{\textbf{User}: Are there any buildings visible at the airport? \\ \textbf{Answer}: No.}} \\
 & RSVQA-HR~\cite{lobry2020rsvqa} & RL &  train/test & 1k/1k \\
 & FloodNet~\cite{rahnemoonfar2021floodnet} & RL &  train & 4k \\
\midrule
\multirow{5}{*}{Image Caption} & VRSBench~\cite{NEURIPS2024_05b7f821} & SFT/RL & train/test & 20k/16k & \multirow{5}{*}{\makecell[l]{\textbf{User}: Describe this image in detail. \\ \textbf{Answer}: The image from GoogleEarth shows a landscape dominated \\ by brownish patches of vegetation with a small baseball field visible in \\ the lower central part of the view. }} \\
 & FIT-RS-cap~\cite{luo2024skysensegpt} & SFT/RL & train & 83k \\
 & NWPU-Captions~\cite{cheng2022nwpu} & RL &  train/test & 25k/6k \\
 & RSICD~\cite{lu2017exploring} & RL &  train/test & 9k/2k \\
 & RSITMD~\cite{yuan2021exploring} & RL &  train/test & 4k/1k \\

\bottomrule
\end{tabular}
}
\caption{Comprehensive overview of datasets and task formulations. We detail the specific data sources, corresponding alignment stages (SFT/RL), sample sizes across train/test splits, and provide representative instruction-response templates for each of the six core remote sensing tasks evaluated in our study.}
\label{tab:supp_dataset_stat}
\vspace{-1.0cm}
\end{table}

\subsection{Datasets and Evaluation Benchmarks}
\label{supp:datasets}

To ensure full transparency and reproducibility of our GeoSolver framework, we provide a comprehensive breakdown of the datasets utilized during the Supervised Fine-Tuning (SFT) and Reinforcement Learning (RL) stages, as well as the evaluation benchmarks. 

As detailed in Table~\ref{tab:supp_dataset_stat}, we construct the alignment data and evaluate GeoSolver across six foundational remote sensing tasks: Visual Grounding, Object Detection, Object Counting, Scene Classification, Visual Question Answering (VQA), and Image Caption. Specifically, the data for the SFT stage is sourced from the Geo-CoT380k dataset, whose constituent source datasets are explicitly listed in the table alongside their respective deployment stages (SFT/RL), train/test split sizes, and concrete instruction-response prompt templates. To rigorously ensure a fair evaluation and prevent any potential data leakage, we conducted a thorough cross-check across all datasets, strictly removing any overlapping or duplicate images between the training sets and the downstream evaluation benchmarks. 

\noindent\textbf{Baselines.}
We comprehensively benchmark GeoSolver against three distinct categories of models to ensure a rigorous and multi-dimensional evaluation: Closed-source Commercial Systems: State-of-the-art proprietary multimodal models, including ChatGPT-5~\cite{openai2025gpt5_blog}, Gemini-2.0-Flash~\cite{comanici2025gemini}, and Claude-3.5-Sonnet~\cite{anthropic2025claude4_opus_sonnet}. General-purpose and Remote Sensing VLMs: Leading open-weight vision-language models and dedicated domain-specific experts, including MiniGPT-v2~\cite{zhu2023minigpt}, Qwen2.5-VL~\cite{Qwen2.5-VL}, GeoChat~\cite{kuckreja2024geochat}, VHM~\cite{pang2025vhm}, SkysenseGPT~\cite{luo2024skysensegpt}, and EarthDial~\cite{soni2025earthdial}. Reasoning-centric Frameworks: Recent models explicitly optimized or prompted for Chain-of-Thought reasoning, including GLM-4.1V-9B-Thinking~\cite{vteam2025glm45vglm41vthinkingversatilemultimodal}, Kimi-VL-A3B-Thinking-2506~\cite{kimiteam2025kimivltechnicalreport}, and RS-EoT-7B~\cite{shao2025asking_geoeot}. 

Furthermore, to validate the specific efficacy of our token-level verifier during Test-Time Scaling (TTS) evaluations, we compare GeoPRM against several alternative reward modeling approaches. This includes advanced mathematical and logical Process Reward Models (URSA-8B-RM~\cite{luo2025ursa} and GraphPRM~\cite{peng2025rewarding_graphprm}), as well as powerful generalist VLMs (Qwen3-VL~\cite{Qwen3-VL} and GLM-4.1V~\cite{vteam2025glm45vglm41vthinkingversatilemultimodal}) prompted to act as step-wise verifiers. This comprehensive comparison rigorously isolates the contribution of our domain-specific process supervision from general multimodal capabilities.

\section{Comprehensive Evaluation}
\label{supp:detailed_results}

Due to space constraints in the main manuscript, we reported the aggregated average performance across multiple datasets. In this section, we provide the comprehensive, fine-grained, per-dataset breakdown of our experimental results across Visual Grounding (Table~\ref{tab:exp_vg}), Object Counting (Table~\ref{tab: exp_oc}), Visual Question Answering (Table~\ref{tab: exp_sc_vqa}), and Image Captioning (Table~\ref{tab: exp_ic}). 

These detailed breakdowns consistently reveal a prominent trend: while state-of-the-art closed-source commercial models (e.g., GPT-4o, Claude-3.5-Sonnet) excel in general domains, they suffer from severe performance degradation in geospatial tasks. Conversely, GeoSolver, empowered by process-level verification, consistently establishes new state-of-the-art benchmarks, particularly in tasks demanding rigorous spatial localization and logical deduction.

\begin{table}[!tb]
\vspace{-0.4cm}
\centering
\resizebox{\linewidth}{!}{
\begin{tabular}{lcccccccccccccccc} 
\toprule
\multirow{2}{*}{\textbf{Method}}
& \multicolumn{3}{c}{\textbf{VRSBench-VG}} 
&  \multicolumn{3}{c}{\textbf{DIOR-RSVG}} 
& \multicolumn{3}{c}{\textbf{RRSIS-D}} 
& \multicolumn{3}{c}{\textbf{RSVG}} 
\\

\cmidrule(lr){2-4} \cmidrule(lr){5-7} \cmidrule(lr){8-10} \cmidrule(lr){11-13} 
& \textbf{@0.5} & \textbf{@0.75} & \textbf{mIoU}
& \textbf{@0.5} & \textbf{@0.75} & \textbf{mIoU}
& \textbf{@0.5} & \textbf{@0.75} & \textbf{mIoU}
& \textbf{@0.5} & \textbf{@0.75} & \textbf{mIoU}

\\
\midrule
\multicolumn{7}{l}{\textcolor{gray}{\textit{Close-source Commercial Vision-Language Models}}} \\
Claude-sonnet-4 & 4.2 & 0.6 & 6.65 & 4.1 & 0.0 & 8.87 & 5.5 & 0.5 & 11.62 & 1.0 & 0.0 & 2.53   \\
Gemini-2.0-flash & 8.9 & 1.6 & 13.52 & 3.3 & 0.8 & 11.40 & 10.5 & 0.0 & 16.05 & 0.0 & 0.0 & 1.75  \\
ChatGPT-5 & 2.4 & 0.2 & 5.65 & 5.3 & 0.0 & 8.27 & 8.5 & 0.5 & 10.60 & 0.0 & 0.0 & 1.49  \\
\midrule

\multicolumn{7}{l}{\textcolor{gray}{\textit{Open-source Reasoning Vision-Language Models}}} \\
GLM-4.1V & \underline{46.0} & \underline{25.3} & \underline{43.20} & 40.0 & 25.3 & 39.41 & 45.0 & 27.5 & 43.03 & \underline{19.5} & \underline{5.5} & \underline{19.35} \\
\midrule 
\multicolumn{7}{l}{\textcolor{gray}{\textit{Open-source Remote Sensing Vision-Language Models}}} \\
GeoChat & 14.2 & 2.1 & 23.66 & 25.9 & 3.3 & 29.87 & 24.4 & 3.2 & 30.30 & 0.0 & 0.0 & 0.13  \\
VHM & 33.9 & 10.0 & 34.91 & \underline{55.9} & \underline{35.5} & \underline{49.90} & \underline{64.0} & \underline{37.5} & \underline{55.20} & 2.5 & 0.0 & 5.80   \\
SkySenseGPT & 13.4 & 2.7 & 21.25 & 22.9 & 2.6 & 27.76 & 27.8 & 8.4 & 33.63 & 1.0 & 0.5 & 4.57  \\
EarthDial  & 14.4 & 7.8 & 13.04 & 46.1 & 30.2 & 39.5 & 34.5 & 15.5 & 35.39 & 1.0 & 0.0 & 7.18   \\
\midrule

\textbf{GeoSolver} & \textbf{78.4} & \textbf{43.5} & \textbf{64.57} & \textbf{87.8} & \textbf{68.2} & \textbf{75.62} & \textbf{86.3} & \textbf{70.7} & \textbf{76.66} & \textbf{26.5} & \textbf{7.0} & \textbf{27.43} \\
\bottomrule
\end{tabular}
}
\caption{Comparison of GeoSolver with existing generic and RS VLMs on Visual Grounding task. }
\label{tab:exp_vg}
\vspace{-0.5cm}
\end{table}

\subsection{Visual Grounding}
As presented in Table~\ref{tab:exp_vg}, Visual Grounding requires the model to output precise bounding box coordinates for a given textual query. Commercial models nearly fail this task (mIoU typically below 15\%), highlighting the massive domain gap in interpreting top-down perspectives and arbitrary orientations. Furthermore, compared to existing open-source remote sensing VLMs (e.g., EarthDial, VHM), GeoSolver demonstrates a formidable advantage, especially under the strict Acc@0.75 accuracy threshold. For instance, on the DIOR-RSVG dataset, GeoSolver achieves an outstanding 68.2\% at Acc@0.75, vastly outperforming VHM (35.5\%) and GLM-4.1V (25.3\%). This validates that our token-level drop-moment penalty effectively prevents the model from hallucinating ungrounded coordinates, ensuring pixel-level faithfulness.

\subsection{Object Counting}
Table~\ref{tab: exp_oc} details the performance on Object Counting, evaluated via Accuracy (Acc, higher is better) and Mean Absolute Error (MAE, lower is better). Counting dense, small objects in satellite imagery is notoriously prone to logical shortcuts and visual omissions. Without process supervision, baseline models often guess the count based on holistic scene context rather than rigorous enumeration. Guided by the verifiable Chain-of-Thought format, GeoSolver achieves the lowest MAE across DOTAv2-val (2.927), HRRSD (0.281), RSOD (1.225), and NWPU-VHR (0.405).

\begin{table}[!tb]
\setlength{\tabcolsep}{4pt}
\resizebox{\linewidth}{!}{
\begin{tabular}{lcccccccc} 
\toprule

\multirow{2}{*}{\textbf{Method}} 
& \multicolumn{2}{c}{\textbf{DOTAv2-val}}
& \multicolumn{2}{c}{\textbf{HRRSD}} 
& \multicolumn{2}{c}{\textbf{RSOD}} 
& \multicolumn{2}{c}{\textbf{NWPU-VHR}} 
\\
\cmidrule(lr){2-3} \cmidrule(lr){4-5} \cmidrule(lr){6-7} \cmidrule(lr){8-9} 
& \textbf{Acc\(\uparrow\)} & \textbf{MAE\(\downarrow\)} 
& \textbf{Acc\(\uparrow\)} & \textbf{MAE\(\downarrow\)} 
& \textbf{Acc\(\uparrow\)} & \textbf{MAE\(\downarrow\)} 
& \textbf{Acc\(\uparrow\)} & \textbf{MAE\(\downarrow\)} 

\\
\midrule
\multicolumn{7}{l}{\textcolor{gray}{\textit{Close-source Commercial Vision-Language Models}}} \\
Claude-sonnet-4 & 25.17 & 10.232 & 50.11 & 2.231 & 25.0 & 4.115 & 51.5 & 2.205  \\
Gemini-2.0-flash & 29.36 & 15.057 & 54.65 & 1.921 & 39.0 & 4.095 & \underline{63.5} & 1.835 \\
ChatGPT-5 & \underline{36.20} & 7.490 & 58.50 & \underline{0.787} & 40.0 & 1.430 & 58.0 & 1.310   \\
\midrule
\multicolumn{7}{l}{\textcolor{gray}{\textit{Open-source Vision-Language Models}}} \\
MiniGPT-v2 & 10.82 & 57.082 & 19.50 & 36.059 & 19.5 & 9.630 & 21.0 & 4.675  \\
Qwen2.5-VL & 33.77 & 9.733 & 57.82 & 0.846 & 42.0 & \underline{1.370} & 58.0 & \underline{1.170}   \\
\midrule
\multicolumn{7}{l}{\textcolor{gray}{\textit{Open-source Reasoning Vision-Language Models}}} \\
Kimi-VL-Thinking & 30.68 & 11.967 & 46.26 & 1.612 & 15.5 & 4.050 & 53.0 & 2.575  \\
GLM-4.1V-Thinking & 29.80 & 8.072 & 58.96 & 0.903 & 28.5 & 3.220 & 62.5 & 1.194  \\

\midrule 
\multicolumn{7}{l}{\textcolor{gray}{\textit{Open-source Remote Sensing Vision-Language Models}}} \\
VHM & 32.67 & 9.260 & 46.71 & 1.063  &16.0 & 1.791  & 48.5 & 1.289  \\
SkySenseGPT & 33.11 & \underline{7.199} & 58.73 & 1.070 & \textbf{51.5} & 3.079  & 49.5 & 1.835  \\
EarthDial & 32.23 & 8.422 & \underline{61.45} & 0.871  & 41.0 & 1.642 & 52.5 & 1.323  \\

\midrule

\textbf{GeoSolver} & \textbf{45.92} & \textbf{2.927} & \textbf{84.13} & \textbf{0.281} & \underline{45.5} & \textbf{1.225} & \textbf{79.0} & \textbf{0.405} \\
\bottomrule
\end{tabular}
}
\caption{Comparison of GeoSolver with existing generic and RS VLMs on Object Counting task. }
\label{tab: exp_oc}
\vspace{-0.6cm}
\end{table}

\subsection{Visual Question Answering and Scene Classification}
In Table~\ref{tab: exp_sc_vqa}, we present a granular analysis of VQA performance on VRSBench-VQA and RSVQA-HR, categorized by specific question types (e.g., Existence, Position, Quantity, Scene). The superiority of GeoSolver is most pronounced in the Position (71.55\%) and Quantity (60.67\%) categories of VRSBench-VQA. These specific question types inherently demand multi-step spatial reasoning rather than global semantic matching. The performance leap over strong baselines like Kimi-VL-Thinking and EarthDial explicitly demonstrates that our Process-Aware Tree-GRPO aligns the model's cognitive process with the complex topological structures unique to remote sensing.

\begin{table}[!tb]
\begin{center}
\resizebox{\linewidth}{!}{
\begin{tabular}{lccccccccc}
\toprule
\multirow{2}{*}{\textbf{Method}} & \multicolumn{7}{c}{\textbf{VRSBench-VQA}} & \multicolumn{2}{c}{\textbf{RSVQA-HR}}  \\
\cmidrule(lr){2-8} \cmidrule(lr){9-10}
& \textbf{Category} & \textbf{Existence} & \textbf{Position} & \textbf{Quantity} & \textbf{Scene} & \textbf{Color} & \textbf{Image} & \textbf{Presence} & \textbf{Comp} \\
\midrule
\multicolumn{7}{l}{\textcolor{gray}{\textit{Close-source Commercial Vision-Language Models}}} \\
Claude-sonnet-4 & 43.28 & 52.78 & 30.17 & \underline{66.67} & 64.79 & \underline{63.29} & \underline{91.67}  & 46.95 & 64.94 \\
Gemini-2.0-flash  & 44.03 & 86.11 & 43.97 & 46.00 & 60.56 & 56.96 & \textbf{95.83}  & 56.94 & 42.96   \\
ChatGPT-5  & 39.55 & \underline{88.89} & 42.24 & 47.33 & 70.42 & 59.49 & 87.50  & 62.94 & 68.93  \\
\midrule
\multicolumn{7}{l}{\textcolor{gray}{\textit{Open-source Vision-Language Models}}} \\
MiniGPT-v2  & 25.37 & 56.25 & 20.69 & 44.00 & 45.07 & 36.71 & 33.33  & 48.95 & 52.95 \\
Qwen2.5-VL & 37.31 & 75.69 & 37.93 & 44.00 & 67.61 & \underline{63.29} & \underline{91.67}  & 57.92 & 56.94 \\
\midrule
\multicolumn{7}{l}{\textcolor{gray}{\textit{Open-source Reasoning Vision-Language Models}}} \\
Kimi-VL-Thinking & 47.01 & 87.50 & \underline{46.55} & \textbf{74.67} & \underline{71.83} & \textbf{65.82} & 90.23  & 63.94 & 77.91 \\
GLM-4.1V-Thinking & 42.54 & 86.11 & 43.10 & 54.67 & 69.01 & 62.03 & 87.50 & 45.95 & 65.93  \\

\midrule 
\multicolumn{7}{l}{\textcolor{gray}{\textit{Open-source Remote Sensing Vision-Language Models}}} \\
VHM  & 50.75 & 86.81 & 36.21 & 42.67 & 53.52 & 55.70 & 54.17  & 61.94 & 76.92 \\
SkySenseGPT & \underline{57.46} & 84.03 & 44.83 & 38.00 & 53.52 & 16.46 & 45.83 &  47.95 & \underline{78.93} \\
EarthDial & 51.49 & 47.22 & 36.21 & 41.33 & 36.62 & 11.39 & 50.00  & \underline{64.94} & \textbf{79.92} \\

\midrule
\textbf{GeoSolver} & \textbf{85.82} & \textbf{95.14} & \textbf{71.55} & 60.67 & \textbf{76.05} & 58.23 & \underline{91.67} & \textbf{67.83} & 78.55 \\
\bottomrule
\end{tabular}
}
\caption{Comparison of GeoSolver with generic and RS VLMs on Classification and VQA tasks. }
\label{tab: exp_sc_vqa}
\end{center}
\vspace{-1.02cm}
\end{table}

\begin{table}[!tb]
\begin{center}
\setlength{\tabcolsep}{4pt}
\begin{minipage}[t]{\linewidth}
\centering
\resizebox{\textwidth}{!}{
\begin{tabular}{lcccccccccccc}

\toprule
\multirow{2}{*}{\textbf{Method}} & \multicolumn{3}{c}{\textbf{RSITMD}} & \multicolumn{3}{c}{\textbf{NWPU-Captions}} & \multicolumn{3}{c}{\textbf{RSICD}} & \multicolumn{3}{c}{\textbf{VRSBench-Cap}} \\
\cmidrule(lr){2-4} \cmidrule(lr){5-7} \cmidrule(lr){8-10} \cmidrule(lr){11-13}

&\textbf{B-4} & \textbf{MT} & \textbf{Cr} 
&\textbf{B-4} & \textbf{MT} & \textbf{Cr} 
&\textbf{B-4} & \textbf{MT} & \textbf{Cr} 
&\textbf{B-4} & \textbf{MT} & \textbf{Cr} 
\\
\midrule
\multicolumn{13}{l}{\textcolor{gray}{\textit{Close-source Commercial Vision-Language Models}}} \\
Claude-sonnet-4 & 20.14 & 17.15 & 19.31 & 28.32 & 21.98 & 32.46 & 11.58 & 13.90 & 24.57 & 14.62 & 22.36 & 73.49 \\
Gemini-2.0-flash & 15.73 & 9.27 & 17.11 & 20.55 & 11.42 & 22.58 & 10.85 & 8.71 & 21.53 & 14.19 & 22.30 & 86.33  \\
ChatGPT-5 & 27.27 & 21.10 & 29.48 & 39.62 & 25.69 & 48.52 & 16.83 & 16.73 & 34.39 & 18.06 & \textbf{25.11} & 88.93  \\
\midrule
\multicolumn{13}{l}{\textcolor{gray}{\textit{Open-source Vision-Language Models}}} \\
MiniGPT-v2\nocite{chenminigpt} & 25.45 & 16.83 & 25.89 & 37.75 & 19.70 & 35.73 & 15.40 & 12.36 & 26.63 & 26.61 & 18.36 & 68.94  \\
Qwen2.5-VL\nocite{bai2025qwen2} & 27.92 & 17.24 & 24.90 & 38.89 & 21.40 & 42.11 & 17.80 & 13.72 & 32.19 & 29.21 & \underline{25.01} & 91.84  \\
\midrule 
\multicolumn{13}{l}{\textcolor{gray}{\textit{Open-source Reasoning Vision-Language Models}}} \\
Kimi-VL-Thinking  & 24.82 & 16.47 & 22.02 & 34.84 & 20.08 & 37.14  & 15.60 & 13.57 & 30.00 & 26.07 & 24.34 & 83.86 \\
GLM-4.1V-Thinking & 20.57 & 19.55 & 24.98 & 29.59 & 23.33 & 40.35 & 12.57 & 15.86 & 30.47 & 13.52 & 22.57 & 79.71 \\
\midrule 
\multicolumn{13}{l}{\textcolor{gray}{\textit{Open-source Remote Sensing Vision-Language Models}}} \\
VHM\nocite{pang2025vhm} &38.93 &21.99 &40.29 &50.69 &25.31 &54.92 &25.66 &17.63 &49.80 & \textbf{35.06} &22.29 &99.82\\
SkySenseGPT\nocite{luo2024skysensegpt} & 37.76 & 19.06 & 34.98 & 23.33 & 14.02 & 40.48 & \textbf{42.47} & 24.95 & 52.58 & 33.10 & 22.50 & \underline{102.8} \\
EarthDial\nocite{soni2025earthdial} & \underline{42.09} & \underline{23.92} & \underline{42.56} & \underline{67.14} & \underline{46.17} & \textbf{123.6} & 29.09 & \underline{25.20} & \underline{85.82} & 21.49 & 15.88 & 90.51 \\
\midrule

\textbf{GeoSolver} & \textbf{52.50} & \textbf{31.66} & \textbf{66.85} & \textbf{80.93} & \textbf{49.80} & \underline{81.78} & \underline{36.18} & \textbf{26.89} & \textbf{97.24} & \textbf{35.44} & 22.24 & \textbf{106.8}   \\
\bottomrule
\end{tabular}
}
\begin{tablenotes}[flushleft]
\item[] \tiny\textit{B-4 / MT / Cr: BLEU-4 / METEOR / CIDEr}
\end{tablenotes}
\end{minipage}
\end{center}
\vspace{-0.25cm}
\caption{Comparison of GeoSolver with existing generic and RS VLMs on Image Captioning task. }
\label{tab: exp_ic}
\vspace{-1.0cm}
\end{table}

\subsection{Image Caption}
Table~\ref{tab: exp_ic} reports the generative capabilities of the models using standard NLG metrics (BLEU-4, METEOR, CIDEr). While Image Caption is primarily a descriptive task, capturing the nuanced spatial relationships and high-density object distributions in remote sensing imagery requires strong intrinsic visual grounding. GeoSolver achieves highly competitive or state-of-the-art performance, recording a remarkable CIDEr score of 106.8 on VRSBench-Cap. This indicates that the rigorous logical alignment acquired during RL not only eliminates hallucinations in discriminative tasks but also translates into richer, more factually accurate generative descriptions.

\begin{figure}[htpb]
    \centering
    \begin{subfigure}{0.32\linewidth}
        \centering
        \includegraphics[width=\linewidth]{test_time_scaling_geosolver_VG.pdf}
        \caption{Visual Grounding}
        \label{fig:tts_sub1}
    \end{subfigure}\hfill
    \begin{subfigure}{0.32\linewidth}
        \centering
        \includegraphics[width=\linewidth]{test_time_scaling_geosolver_VQA.pdf}
        \caption{VQA}
        \label{fig:tts_sub2}
    \end{subfigure}\hfill
    \begin{subfigure}{0.32\linewidth}
        \centering
        \includegraphics[width=\linewidth]{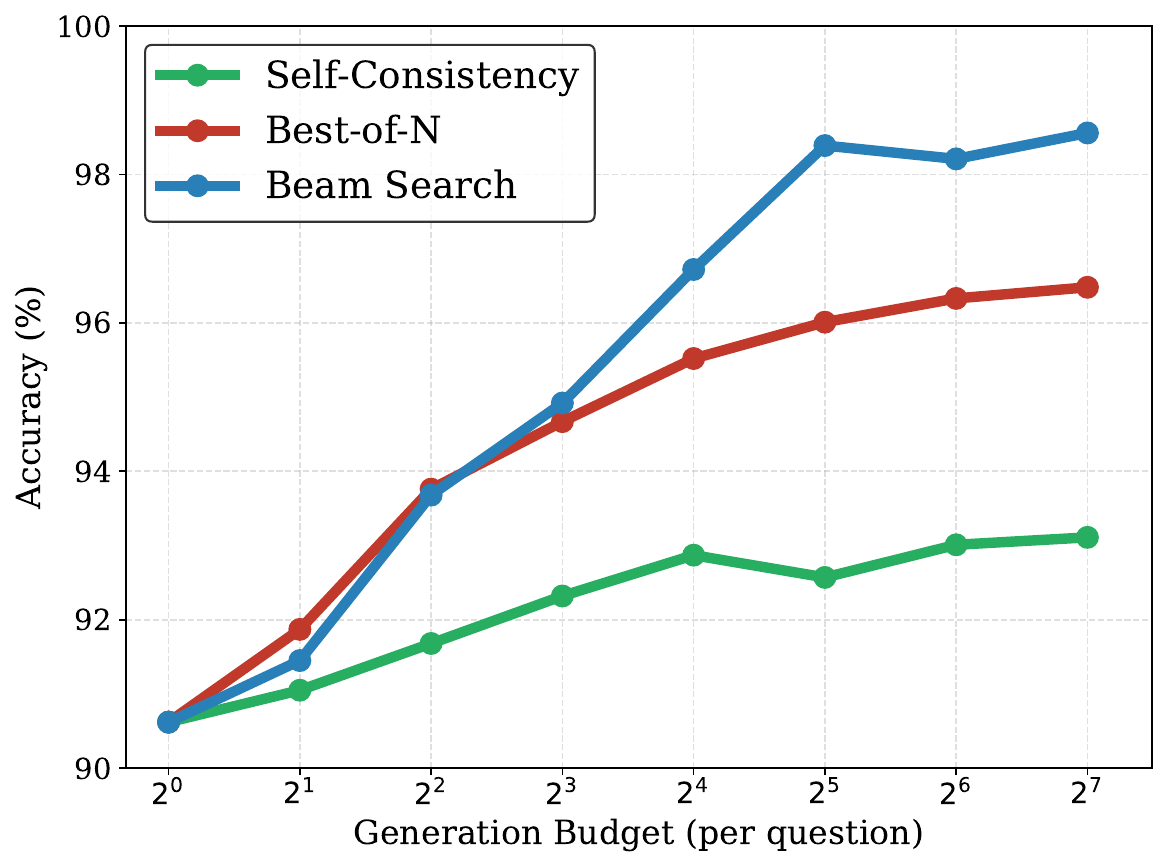}
        \caption{Scene Classification}
        \label{fig:tts_sub3}
    \end{subfigure}
    
    \vspace{0.2cm} 
    
    \begin{subfigure}{0.32\linewidth}
        \centering
        \includegraphics[width=\linewidth]{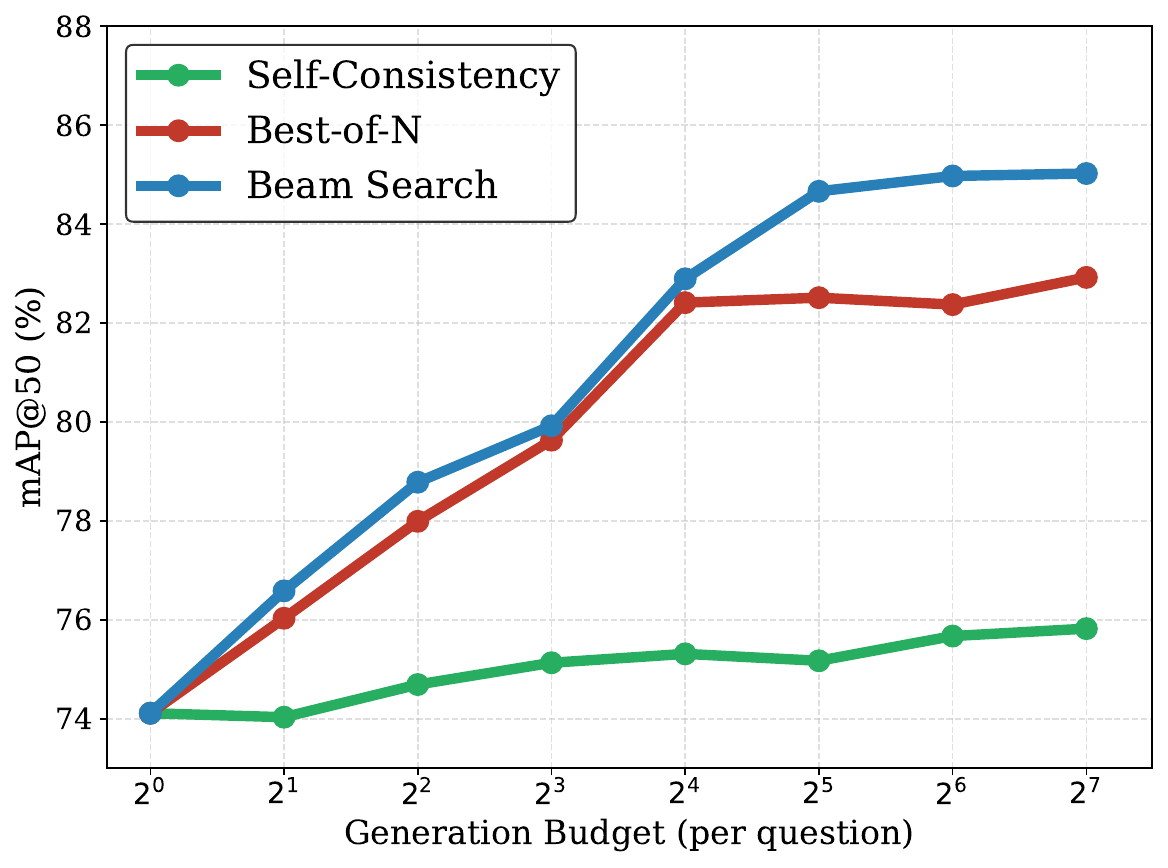}
        \caption{Object Detecting}
        \label{fig:tts_sub4}
    \end{subfigure}\hfill
    \begin{subfigure}{0.32\linewidth}
        \centering
        \includegraphics[width=\linewidth]{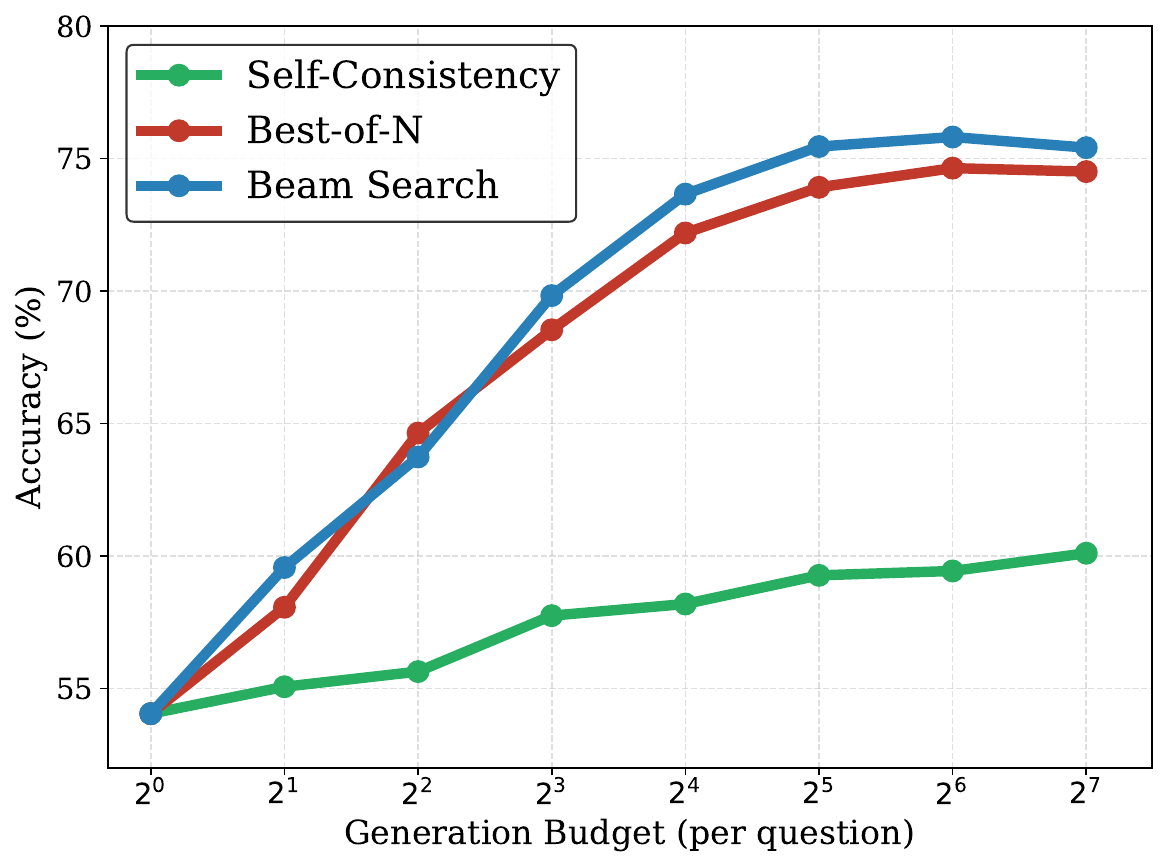}
        \caption{Object Couting}
        \label{fig:tts_sub5}
    \end{subfigure}\hfill
    \begin{subfigure}{0.32\linewidth}
        \centering
        \includegraphics[width=\linewidth]{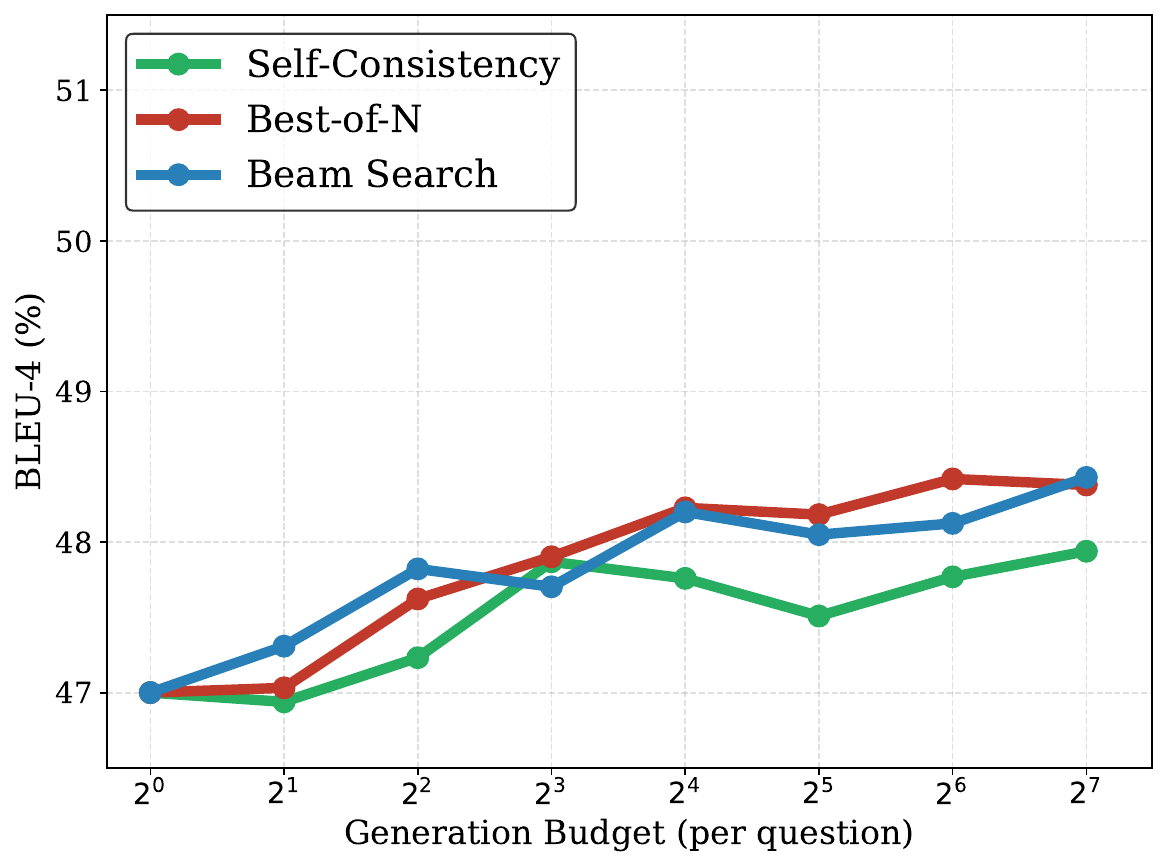}
        \caption{Image Caption}
        \label{fig:tts_sub6}
    \end{subfigure}
    
   \caption{Comprehensive Test-Time Scaling (TTS) curves across all six remote sensing tasks. We compare the scaling behavior of Self-Consistency (Green), Best-of-N (Red), and Beam Search (Blue) as the inference generation budget ($N$) scales from $2^0$ to $2^7$. Guided by our token-level GeoPRM, both search strategies unlock robust compute-optimal scaling, significantly and consistently outperforming the unverified Self-Consistency baseline.}
    \label{fig:tts_6_images}
    \vspace{-0.6cm}
\end{figure}

\subsection{Complete Test-Time Scaling Curves}
\label{supp:full_tts_curves}

In the main manuscript, due to space limitations, we only illustrated the Test-Time Scaling (TTS) behavior for Visual Grounding and VQA. To provide a comprehensive view of GeoPRM's verification capabilities, Figure~\ref{fig:tts_6_images} presents the complete scaling curves across all six evaluated remote sensing tasks as the generation budget ($N$) increases exponentially from $2^0$ to $2^7$. These extensive results robustly corroborate the conclusions drawn in our main text, revealing several consistent trends across diverse reasoning paradigms.

\noindent\textbf{Saturation of Unverified Sampling.} 
As observed in the curves (green lines), Self-Consistency (majority voting) provides only marginal early gains before quickly plateauing across almost all tasks. Because it fundamentally relies on the unverified sampling distribution of the base policy, it is incapable of correcting intrinsic logical flaws or severe spatial hallucinations, regardless of how much inference compute is allocated.

\noindent\textbf{Compute-Optimal Scaling via GeoPRM.} 
In stark contrast, integrating GeoPRM as a process verifier unlocks explicit scaling laws. Both Best-of-N (red lines) and Beam Search (blue lines) exhibit continuous, substantial performance gains as the generation budget increases. This demonstrates that GeoPRM's token-level verification successfully identifies and rewards faithful reasoning trajectories that are otherwise buried in the base model's default output distribution.

\noindent\textbf{Superiority of Beam Search in Complex Logic.} 
Consistent with our primary analysis, Beam Search shows a distinct advantage over Best-of-N, particularly at higher compute budgets and in tasks demanding rigorous topological reasoning, such as Object Detection (Figure~\ref{fig:tts_sub4}) and Scene Classification (Figure~\ref{fig:tts_sub3}). By leveraging GeoPRM's drop-moment penalty to prune hallucinatory branches step-by-step during autoregressive generation, Beam Search explores the optimal solution space much more efficiently than post-hoc sequence-level filtering.

\noindent\textbf{Generative Task Behavior.} 
For generative tasks like Image Captioning (Figure~\ref{fig:tts_sub6}), while the N-gram based metric (BLEU-4) inherently exhibits more variance, GeoPRM-guided scaling still maintains a clear upward trajectory compared to the stagnant baseline. This proves the effectiveness of rigorous process supervision even beyond strict discriminative scenarios.

\begin{figure}[htpb]
    \vspace{-0.3cm}
    \centering
    \begin{subfigure}{0.32\linewidth}
        \centering
        \includegraphics[width=\linewidth]{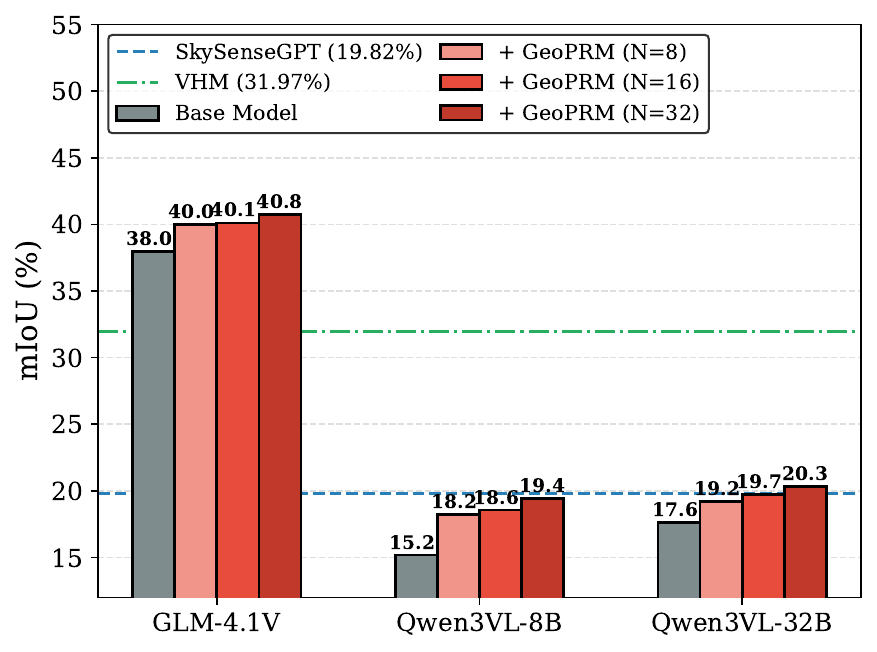}
        \caption{Visual Grounding}
        \label{fig:gener_sub1}
    \end{subfigure}\hfill
    \begin{subfigure}{0.32\linewidth}
        \centering
        \includegraphics[width=\linewidth]{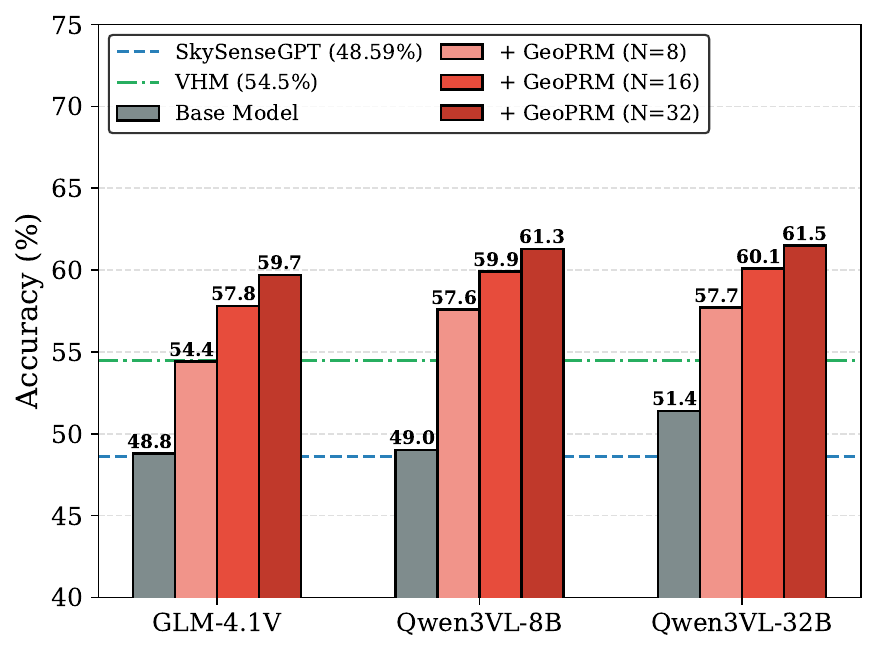}
        \caption{VQA}
        \label{fig:gener_sub2}
    \end{subfigure}\hfill
    \begin{subfigure}{0.32\linewidth}
        \centering
        \includegraphics[width=\linewidth]{generalization_geoprm_SC_half.pdf}
        \caption{Scene Classification}
        \label{fig:gener_sub3}
    \end{subfigure}
    
    \vspace{0.2cm} 
    
    \begin{subfigure}{0.32\linewidth}
        \centering
        \includegraphics[width=\linewidth]{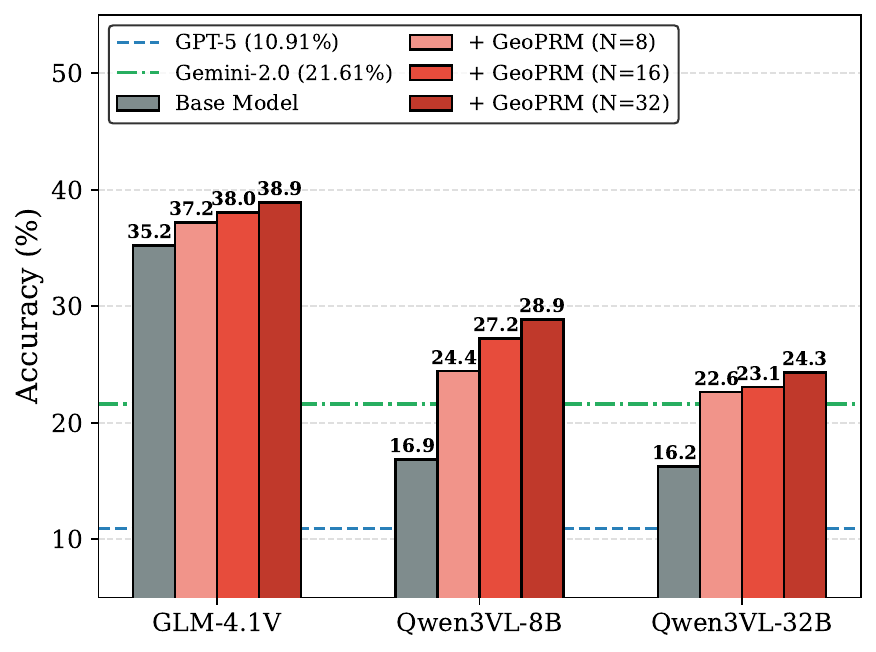}
        \caption{Object Detection}
        \label{fig:gener_sub4}
    \end{subfigure}\hfill
    \begin{subfigure}{0.32\linewidth}
        \centering
        \includegraphics[width=\linewidth]{generalization_geoprm_OC_half.pdf}
        \caption{Object Counting}
        \label{fig:gener_sub5}
    \end{subfigure}\hfill
    \begin{subfigure}{0.32\linewidth}
        \centering
        \includegraphics[width=\linewidth]{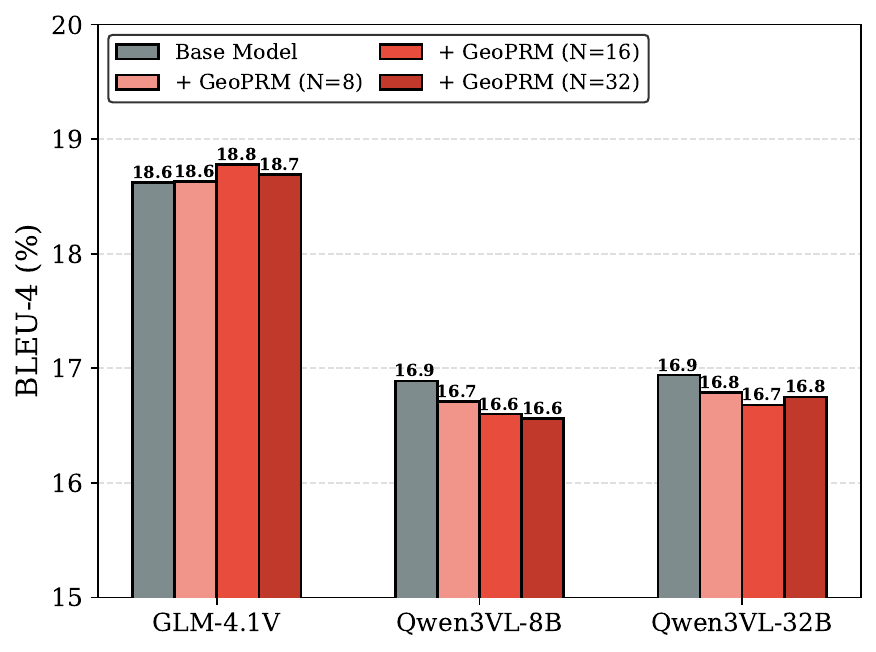}
        \caption{Image Caption}
        \label{fig:gener_sub6}
    \end{subfigure}
    \vspace{-0.3cm}
    \caption{Comprehensive cross-model generalization evaluation across six remote sensing tasks. Gray bars indicate the greedy decoding performance of the base generalist VLMs (GLM-4.1V-9B, Qwen3VL-8B, Qwen3VL-32B). The red gradient bars demonstrate the progressive performance gains achieved by incorporating GeoPRM as a verifier under Best-of-N ($N \in \{8, 16, 32\}$) decoding. Horizontal dashed and dotted lines represent the baseline performance of state-of-the-art, fully fine-tuned domain experts (e.g., SkySenseGPT, VHM, EarthDial). Guided by GeoPRM, general-purpose models successfully bridge the domain gap and frequently surpass dedicated remote sensing specialists.}
    \label{fig:gener_6_images}
    \vspace{-0.8cm}
\end{figure}

\subsection{Complete Cross-Model Generalization Results}
\label{supp:cross_model_full}

In the main manuscript, we demonstrated GeoPRM's capability to generalize as a universal verifier by presenting its Test-Time Scaling performance on Object Counting and Scene Classification. To fully substantiate this claim, Figure~\ref{fig:gener_6_images} visualizes the complete cross-model generalization results across all six geospatial reasoning tasks. We evaluate the zero-shot base performance and the GeoPRM-guided BoN performance ($N \in \{8, 16, 32\}$) on three distinct general-purpose VLMs: GLM-4.1V-9B~\cite{vteam2025glm45vglm41vthinkingversatilemultimodal}, Qwen3VL-8B~\cite{Qwen3-VL}, and Qwen3VL-32B~\cite{Qwen3-VL}.

\noindent\textbf{Consistent Verification Across Architectures.} 
The results reveal a remarkably consistent trend across virtually all tasks: equipping general-purpose models with GeoPRM yields monotonic performance improvements as the generation budget $N$ increases. This consistency across different base model architectures (GLM vs. Qwen) and model scales (8B vs. 32B) confirms that GeoPRM has not simply overfit to the GeoSolver policy. Instead, it successfully captures a fundamental, transferable logic for evaluating multimodal geospatial reasoning.

\noindent\textbf{Surpassing Fully Fine-Tuned Domain Experts.} 
The most compelling observation emerges when comparing the scaled generalist models against dedicated remote sensing experts (represented by horizontal reference lines). In highly complex spatial tasks such as Visual Grounding (Figure~\ref{fig:gener_6_images}a), VQA (Figure~\ref{fig:gener_6_images}b), and Object Detection (Figure~\ref{fig:gener_6_images}d), the base generalist models initially lag behind domain experts like VHM or SkySenseGPT. However, guided by GeoPRM at $N=32$, these generalists not only bridge this domain gap but frequently surpass the fine-tuned experts. This highlights the profound inefficiency of relying solely on outcome-supervised SFT, and demonstrates that process-level verification is a far more robust paradigm for unlocking spatial intelligence.

\end{document}